\documentclass[conference]{IEEEtran}
\IEEEoverridecommandlockouts
\usepackage{cite}
\usepackage{amsmath,amssymb,amsfonts}
\usepackage{algorithm} 
\usepackage{algpseudocode}
\usepackage{booktabs}
\usepackage{graphicx}
\usepackage{textcomp}
\usepackage{times}
\usepackage{ragged2e}
\usepackage{blindtext}
\usepackage{hyperref}
\usepackage{amsmath}
\usepackage{longtable}
\usepackage{footnote}
\usepackage{xurl}
\usepackage{comment}
\usepackage[T1]{fontenc}
\usepackage{hyphenat}
\usepackage[flushleft]{threeparttable}
\usepackage{tablefootnote}
\usepackage{multirow}
\usepackage{arydshln}
\usepackage{xcolor}
\usepackage{bibunits}

\def\BibTeX{{\rm B\kern-.05em{\sc i\kern-.025em b}\kern-.08em
    T\kern-.1667em\lower.7ex\hbox{E}\kern-.125emX}}

\begin{document}

\title{AI Hallucinations: A Misnomer Worth Clarifying}

\author{\IEEEauthorblockN{Negar Maleki}
\IEEEauthorblockA{\textit{University of South Florida}\\
negarmaleki@usf.edu}
\and
\IEEEauthorblockN{Balaji Padmanabhan}
\IEEEauthorblockA{\textit{University of Maryland}\\
bpadmana@umd.edu}
\and
\IEEEauthorblockN{Kaushik Dutta}
\IEEEauthorblockA{\textit{University of South Florida}\\
duttak@usf.edu}
}

\maketitle

\begin{abstract}
As large language models continue to advance in Artificial Intelligence (AI), text generation systems have been shown to suffer from a problematic phenomenon termed often as “hallucination.” However, with AI's increasing presence across various domains including medicine, concerns have arisen regarding the use of the term itself. In this study, we conducted a systematic review to identify papers defining “AI hallucination” across fourteen databases. We present and analyze definitions obtained across all databases, categorize them based on their applications, and extract key points within each category. Our results highlight a lack of consistency in how the term is used, but also help identify several alternative terms in the literature. We discuss implications of these and call for a more unified effort to bring consistency to an important contemporary AI issue that can affect multiple domains significantly\footnote{© 20xx IEEE. Personal use of this material is permitted. Permission from IEEE must be obtained for all other uses, in any current or future media, including reprinting/republishing this material for advertising or promotional purposes, creating new collective works, for resale or redistribution to servers or lists, or reuse of any copyrighted component of this work in other works.}.
\end{abstract}


\section{Introduction}
One of the early uses of the term "hallucination" in the field of Artificial Intelligence (AI) was in computer vision, in 2000 \cite{840616}, where it was associated with constructive implications such as super-resolution \cite{840616}, image inpainting \cite{xiang2023deep}, and image synthesis \cite{pumarola2018unsupervised}. Interestingly, in this context hallucination was regarded as a valuable asset in computer vision rather than an issue to be circumvented. For instance, an image with low resolution might have been rendered more useful with careful hallucination \cite{840616} that generated additional pixels specifically for this purpose.

Despite this (more positive) beginning, recent research has started to employ the term "hallucination" to describe a specific type of error in image captioning \cite{biten2022let} and adversarial attack in object detection \cite{braunegg2020apricot}. In this context, "hallucination" refers to instances where non-existent objects are erroneously detected or incorrectly localized at their anticipated positions. This latter (more negative) interpretation of "hallucination" in computer vision mirrors its analogous usage in language models. For instance, in 2017, researchers highlighted challenges in language models, such as \textit{"the output of the Neural Machine Translation (NMT) system is often quite fluent but entirely unrelated to the input"} \cite{koehn2017challenges}, or \textit{"language models presume likelihood, but the generated content is ultimately incorrect and unsupported by any information"} \cite{wiseman2017challenges}, which is interpreted as a form of hallucination in AI.

To date, a precise and universally accepted definition of "hallucination" remains absent in the discussions related to this in the increasingly broader field of AI \cite{filippova2020controlled}. Diverse definitions, or implied interpretations, persist; sometimes even contradictory, as previously highlighted within the field of computer vision where multiple, disparate interpretations coexist under the same term. 

Beyond the AI context, and specifically in the medical domain, the term "hallucination" is a psychological concept denoting a specific form of sensory experience \cite{insel2010rethinking}. Ji et al. \cite{ji2023survey}, from the computer science perspective (in ACM Computing Surveys), rationalized the use of the term "hallucination" as \textit{"an unreal perception that feels real"} by drawing from Blom's definition --- \textit{"a percept, experienced by a waking individual, in the absence of an appropriate stimulus from the extracorporeal world."} On the other hand, Østergaard et al. \cite{ostergaard2023false}, from the medical perspective (in Schizophrenia Bulletin, one of the leading journals in the discipline), raised critical concerns regarding even the adoption of the "hallucination" terminology in AI for two primary reasons: 1) The "hallucination" metaphor in AI from this perspective is a misnomer, as AI lacks sensory perceptions, and errors arise from data and prompts rather than the absence of stimuli, and 2) this metaphor is highly stigmatizing, as it associates negative issues in AI with a specific issue in mental illness, particularly schizophrenia, thereby possibly undermining many efforts to reduce stigma in psychiatry and mental health.

Given AI's increasing presence across various domains, including the medical field, concerns have arisen regarding the multifaceted, possibly inappropriate and potentially even harmful use of the term "hallucination" \cite{emsley2023chatgpt, ostergaard2023false}. To address this issue effectively, two potential paths of work offer some promise: 1) The establishment of a consistent and universally applicable terminologies that can be uniformly adopted across all AI-impacted domains will help, particularly if such terminologies lead to the use of more specific and nuanced terms that actually describe the issues they highlight (as we will show later, such vocabulary does exist, but needs more consistent use) and 2) The formulation of a robust and formal definition of "AI hallucination" within the context of AI. These measures are essential to promote clarity and coherence in discussions and research related to "hallucination" in AI, and to mitigate potential confusion and ambiguity in cross-disciplinary applications.

Motivated by these issues, in this paper, we conduct a systematic review of the use of "AI hallucination" across 14 databases with a focus on identifying various definitions that have been used in the literature so far (our review covers more fields than just healthcare and computer science, including  ethical and legal settings, and domains as diverse as physics, sports, etc. in order to explore any broader issues). Recently, two papers (\cite{ji2023survey, ye2023cognitive}) explored the concept of hallucination in Natural Language Generation- (NLG-) specific tasks (e.g., text translation, text summarization, knowledge graph, etc.). Our work builds on these studies to also consider the application of NLG in diverse domains. The pervasive nature of AI extends beyond these specific tasks, affecting numerous domains and applications. Consequently, our broader review done here reveals the extensive utilization of Large Language Models (LLMs) across almost a much broader space of domains to date, and provides a comprehensive understanding of how the term has been leveraged across various fields. Generally we see that research attempting to define "AI hallucination" does so based on their individual understanding and the challenges encountered within their respective fields. The findings from our systematic and broad review underscore the challenge that the term "AI hallucination" lacks a precise, universally accepted definition, resulting in the observation of various characteristics associated with this term across different applications. We present a summary of these different interpretations and provide some guidance going forward.

\section{Methodology}
\begin{table*}[h!]
  \centering
  \scriptsize
  \caption{Method summary}
  \begin{threeparttable}
        \begin{tabular}
    {p{6.75em}lp{15em}lp{25.75em}cccp{4.75em}}
    \toprule
    \textbf{Source/ Database} &       & \Centering{\textbf{Search Category}} &       & \Centering{\textbf{Query Terms}} &       & \multicolumn{1}{p{3.835em}}{\Centering{\textbf{Num. of Papers}}} &       & \Centering{\textbf{Study End Date}\tnote{*}} \\
\cmidrule{1-1}\cmidrule{3-3}\cmidrule{5-5}\cmidrule{7-7}\cmidrule{9-9}    PubMed &       & \Centering{All Field} &       & \Centering{"Artificial Intelligence" AND "Hallucination"} &       & 103   &       & \Centering{09/27/2023} \\
\cmidrule{1-1}\cmidrule{3-3}\cmidrule{5-5}\cmidrule{7-7}\cmidrule{9-9}    MEDLINE &       & \Centering{All Field} &       & \Centering{"AI hallucination"} &       & 157   &       & \Centering{09/28/2023} \\
\cmidrule{1-1}\cmidrule{3-3}\cmidrule{5-5}\cmidrule{7-7}\cmidrule{9-9}    Scopus &       & \Centering{Title, Abstract, or Introduction} &       & \Centering{"AI+hallucination" OR "AI hallucination"} &       & 483   &       & \Centering{09/27/2023} \\
\cmidrule{1-1}\cmidrule{3-3}\cmidrule{5-5}\cmidrule{7-7}\cmidrule{9-9}    \raggedright{PubMed Central} &       & \Centering{Text Word} &       & \Centering{"Artificial Intelligence" AND "Hallucination"} &       & 371   &       & \Centering{09/28/2023} \\
\cmidrule{1-1}\cmidrule{3-3}\cmidrule{5-5}\cmidrule{7-7}\cmidrule{9-9}    Web of Science &       & \Centering{All Field} &       & \Centering{"AI hallucination"} &       & 139   &       & \Centering{09/28/2023} \\
\cmidrule{1-1}\cmidrule{3-3}\cmidrule{5-5}\cmidrule{7-7}\cmidrule{9-9}    \raggedright{BioMed Central} &       & \Centering{All Field} &       & \Centering{"AI hallucination"}  &       & 76    &       & \Centering{09/29/2023} \\
\cmidrule{1-1}\cmidrule{3-3}\cmidrule{5-5}\cmidrule{7-7}\cmidrule{9-9}    Embase &       & \Centering{All Field} &       & \Centering{"AI hallucination"} &       & 80    &       & \Centering{09/29/2023} \\
\cmidrule{1-1}\cmidrule{3-3}\cmidrule{5-5}\cmidrule{7-7}\cmidrule{9-9}    PLOS  &       & \Centering{Body}  &       & \Centering{"AI hallucination"} &       & 885   &       & \Centering{09/29/2023} \\
\cmidrule{1-1}\cmidrule{3-3}\cmidrule{5-5}\cmidrule{7-7}\cmidrule{9-9}    CINAHL &       & \Centering{All Field} &       & \Centering{"Hallucination" AND ("AI" OR "Artificial Intelligence")} &       & 34    &       & \Centering{09/29/2023} \\
\cmidrule{1-1}\cmidrule{3-3}\cmidrule{5-5}\cmidrule{7-7}\cmidrule{9-9}    ACM   &       & \Centering{Full Text}  &       & \Centering{"AI" AND "hallucination"}  &       & 264   &       & \Centering{09/30/2023} \\
\cmidrule{1-1}\cmidrule{3-3}\cmidrule{5-5}\cmidrule{7-7}\cmidrule{9-9}    IEEEXplore &       & \Centering{Full Text}  &       & \Centering{"AI" AND "hallucination"}  &       & 257   &       & \Centering{09/30/2023} \\
\cmidrule{1-1}\cmidrule{3-3}\cmidrule{5-5}\cmidrule{7-7}\cmidrule{9-9}    ScienceDirect &       & \Centering{All Field} &       & \Centering{"AI hallucination" OR ("AI" AND "hallucination")} &       & 769   &       & \Centering{09/30/2023} \\
\cmidrule{1-1}\cmidrule{3-3}\cmidrule{5-5}\cmidrule{7-7}\cmidrule{9-9}    Google Scholar &       & \Centering{All Field} &       & \Centering{"AI hallucination" AND "hallucination in AI"} &       & 89    &       & \Centering{10/01/2023} \\
\cmidrule{1-1}\cmidrule{3-3}\cmidrule{5-5}\cmidrule{7-7}\cmidrule{9-9}    arXiv &       & \Centering{All Field} &       & \Centering{"AI" AND "hallucination"}  &       & 40    &       & \Centering{10/01/2023} \\
    \bottomrule
    \end{tabular}%
          \begin{tablenotes}[normal,flushleft]
           \item [*] The start date is the same for all databases: 01/01/2013 (Date format: mm/dd/yyyy).
      \end{tablenotes}
  \end{threeparttable}
  \label{method}%
\end{table*}%

Our systematic review covered an extensive database search across various domains, including computer science and health, with a focus on the following databases: PubMed, MEDLINE, Scopus, PubMed Central, Web of Science, BioMed Central, Embase, PLOS, CINAHL, ACM, IEEEXplore, ScienceDirect, Google Scholar, and arXiv (no relevant documents were found in MedlinePlus, Cochrane Library, and UpToDate databases, so those were excluded).

Our search methodology was tailored to adapt to the volume of results as well as the relevance of the papers to our research objectives. We manually reviewed every paper that made it through this process in order to identify possible definitions/usage of the term "hallucination" in AI. Given this goal we had to adapt the search in some cases to identify papers most closely relevant to this objective as noted further below. Also, given differences in how search queries are interpreted across the different databases we had to iteratively modify the search term within each database as well in many cases. For clarity, we present all the specific details of this below (in order to be transparent about how we created the subset of papers from which to examine the definitions). However, the summary of these is provided in Table \ref{method}, including details of the study period for each database.

In the PubMed Database, we initiated an advanced search employing the keywords "Artificial Intelligence" AND "Hallucination" within the "All field" category, yielding 103 papers within the last 10 years. However, the query "AI+hallucination" yielded only 3 papers. Conversely, within the Scopus database, searches for "AI hallucination" or "AI+hallucination" resulted in a total of 1445 records across all fields over the same 10-year period. To manage this extensive dataset, we refined our search criteria to focus on the Title, Abstract, or Introduction, which reduced the results to 483 relevant records. A detailed review of each abstract led us to download papers that appeared pertinent to AI hallucination. This approach significantly differed from searching within abstracts alone, which produced only 49 records and missed some relevant documents.

In PubMed Central (PMC), the query "AI+hallucination" yielded just 1 paper. Consequently, we conducted an advanced search using the keywords "Artificial Intelligence" AND "Hallucination" within the "Text Word" category, uncovering 371 records from the past decade. PMC does not provide abstracts, necessitating the manual examination of each paper to assess its relevance to AI hallucination.

In both the MEDLINE and Web of Science databases, we employed the term "AI hallucination" within the "All field" category, yielding 157 and 139 records, respectively, spanning the last 10 years. In the case of MEDLINE, each record underwent a thorough review, and records containing definitions for AI hallucination were downloaded. Conversely, Web of Science offered abstracts for the records, enabling us to screen them individually and select those relevant to AI hallucination.

Within the BioMed Central (BMC) database, a search for "AI hallucination" led to the retrieval of 76 papers published within the last 10 years. Subsequently, we accessed each paper individually and downloaded those containing pertinent definitions of AI hallucination. In the Embase database, our search for "AI hallucination" produced 80 records published within the last 10 years. These records underwent meticulous abstract review, and those relevant to AI hallucination were selectively downloaded for further analysis.

In the PLOS database, our initial search with "AI hallucination" resulted in a substantial 1064 records across all fields for the past 10 years. Given this large dataset, we refined our search to focus on the "Body" section, yielding 885 records. We proceeded to review the abstracts of each paper and downloaded those demonstrating relevance to AI hallucination. Within the CINAHL database, we conducted an advanced search utilizing the terms "AI" or "Artificial Intelligence" combined with "Hallucination" within the "All field" category, yielding 34 records published in the past 10 years.

In the ACM and IEEEXplore databases, we performed advanced searches using the terms "AI" AND "hallucination" within the "Full Text" category, resulting in 264 and 257 records, respectively, spanning the past 10 years. Each record underwent individual review, and those containing definitions related to AI hallucination, particularly within the field of LLMs, were downloaded.

In the ScienceDirect database, searches for "AI hallucination" or ("AI" AND "hallucination") yielded an identical number of records, specifically 769 English records, spanning the last decade. Each record underwent meticulous examination, and we selectively downloaded papers containing defined concepts of AI hallucination within the LLMs domain.

Searching for "AI hallucination" on Google Scholar yielded 17,000 records from the last 10 years, rendering a comprehensive review unfeasible. To address this challenge, we employed Google Scholar's advanced search feature, identifying records containing the exact phrases "AI hallucination" and "hallucination in AI," which reduced the results to 89 records from the last decade. Subsequently, we conducted a meticulous screening of these records to identify those providing definitions for AI hallucination.

Similarly, within the arXiv database, we conducted an advanced search using the keywords "AI" AND "hallucination" within the "All field" category, resulting in the retrieval of 40 relevant papers. As with our prior search, we meticulously examined each paper and downloaded those containing definitions for AI hallucination.

The eligibility criteria encompassed any type of published scientific research or preprints, such as articles, reviews, communications, editorials, and opinions, that contained the following search terms: "AI hallucination," "AI" AND "hallucination," "Hallucination in AI," or ("AI" OR "Artificial Intelligence") AND "hallucination" in any part of the document, including the title, abstract, and full text. As explained for each database, we employed the most appropriate search terms.

Initially, our search yielded 3753 records, in total, matching these criteria. However, we refined our search to focus exclusively on records that offered a definition of "AI hallucination" within the context of LLMs. It is essential to clarify that we excluded other types of hallucination, such as face hallucination, auditory voice/verbal hallucination, etc., as they were not the primary focus of this review. Our search involved thorough examination of entire documents, and we collected any documents that indicated the presence of a definition for AI hallucination.

Our exclusion criterion was limited to non-English records. The precise database search strategy encompassed all available documents from January 1$^{st}$, 2013, to October 1$^{st}$, 2023. In total, we identified 333 records that provide a definition either independently or by inference from a referenced paper. The summary of the methodology is provided in Table \ref{method}, including details of the study period for each database. While our review of this work is one of the broadest to date, we acknowledge limitations that are implicit in the methodology above - particularly ones where we had to reduce the retrieved number of papers to focus on potentially more relevant ones due to the manual nature of our review (i.e. where we individually reviewed each paper to identify how the term was used and extract the relevant definition in the proper context). Therefore, while the definitions presented here are certainly those that were used it is possible we may have missed a few other definitions that may have newer connotations not identified in our work here. All the 333 definitions are provided in the Appendix.

\section{Result}

\begin{table*}[h!]
  \scriptsize
  \centering
  \caption{Alternative terms used}
  \begin{tabular}{p{12em}lp{49em}lp{4.25em}}
    \toprule
    \textbf{Alternative Terms} &       & \multicolumn{1}{l}{\textbf{Definitions}} &       & \textbf{References}\\
    \cmidrule{1-1}\cmidrule{3-3}\cmidrule{5-5}    \multirow{2}[2]{*}{Confabulation} &       & \multicolumn{1}{l}{AI generated responses that sound plausible but are, in fact, incorrect.} &       & \cite{karakas4475000leveraging} \\ 
    \cdashline{3-3}\cdashline{5-5}
    \multicolumn{1}{l}{} &       & \multicolumn{1}{l}{Definition was not provided.} &       & \cite{rodgers2023artificial}  \\
    \cmidrule{1-1}\cmidrule{3-3}\cmidrule{5-5}    Delusion &       & \multicolumn{1}{l}{AI generated responses that are false.} &       & \cite{madden2023assessing}  \\
    \cmidrule{1-1}\cmidrule{3-3}\cmidrule{5-5}    \multirow{3}[2]{*}{Stochastic Parroting} &       &  The repetition of training data or its patterns, rather than actual understanding or reasoning. &       & \cite{li2023dark} \\ \cdashline{3-3} \cdashline{5-5}
    \multicolumn{1}{c}{} &       & LLM model generates confident, specific, and fluent answers that are factually completely wrong. &       & \cite{ge2023} \\ \cdashline{3-3} \cdashline{5-5}
    \multicolumn{1}{c}{} &       & Definition was not provided. &       & \cite{curtis2023chatgpt} \\
    \cmidrule{1-1}\cmidrule{3-3}\cmidrule{5-5}    Factual Errors &       & Inaccuracies in information or statements that are not in accordance with reality or the truth, often unintentional but resulting in incorrect or misleading information. &       & \cite{borji2023categorical}  \\
    \cmidrule{1-1}\cmidrule{3-3}\cmidrule{5-5}    Fact Fabrication  &       & The occurrence where inaccurate information is invented, not represented in the training dataset, and is presented lucidly. &       & \cite{thirunavukarasu2023large} \\
    \cmidrule{1-1}\cmidrule{3-3}\cmidrule{5-5}    \multirow{3}[2]{*}{Fabrication} &       & The phenomenon where, as a generative AI, ChatGPT generates outputs based on statistical prediction of the text without human-like reasoning, potentially resulting in plausible-sounding but inaccurate responses. &       & \cite{ting2023chatgpt} \\ \cdashline{3-3} \cdashline{5-5}
    \multicolumn{1}{l}{} &       & The phenomenon in ChatGPT output where the text is cogent but not necessarily true. &       & \cite{sriwastwa2023generative} \\ \cdashline{3-3} \cdashline{5-5}
    \multicolumn{1}{l}{} &       & Definition was not provided. &       & \cite{gravel2023learning} \\
    \cmidrule{1-1}\cmidrule{3-3}\cmidrule{5-5}    \raggedright{Falsification and Fabrication} &       & \multicolumn{1}{l}{Definition was not provided.} &       & \cite{emsley2023chatgpt} \\
    \cmidrule{1-1}\cmidrule{3-3}\cmidrule{5-5}    Mistakes, Blunders, Falsehoods &       & \multicolumn{1}{l}{Answers that are fabricated when data are insufficient for an accurate response.} &       & \cite{bryant2023} \\
    \cmidrule{1-1}\cmidrule{3-3}\cmidrule{5-5}    \raggedright{Hasty Generalizations, False Analogy, False Dilemma} &       & AI models making inferences that do not follow from the premises; also “hasty generalizations,” i.e., the fallacy of making (too) strong claims based on (too) limited data. &       & \cite{ostergaard2023false} \\
        \bottomrule
    \end{tabular}%
  \label{terms}%
\vspace{0.3in}
\end{table*}%

We reviewed all retrieved papers and documented the definitions provided in each. One main takeaway was that a formal and consistent definition of hallucination simply does not currently exist. There is also little agreement on the specific characteristics of AI hallucination. Depending on the application, we observe varying characteristics, sometimes even contradictory ones.

For instance, in the context of text translation, Koehn and Knowles \cite{koehn2017challenges} described hallucination as "fluent but irrelevant," or Miao et al. \cite{miao2021prevent} characterized it as "fluent but inadequate," while Lee et al. \cite{lee2019hallucinations} attributed "abnormal and unrelated" characteristics to it, thus illustrating different attributes within the same context. In the text summarization context, hallucination refers to generated content that is inconsistent with the source document \cite{zhao2020reducing, dong2020multi}, with some studies categorizing it into subtypes: "Intrinsic hallucination" and "Extrinsic hallucination" \cite{maynez2020faithfulness, lyu2022faithful}, raising concerns, particularly regarding the latter.

Before the launch of ChatGPT on November 30, 2022, we hardly observed definitions for AI hallucination in fields other than computer science. However, with the advent of ChatGPT, researchers have recognized the urgent need for Large Language Models (LLMs) in various fields, including medicine. Therefore, over time, we have observed that the definition has changed and seems to have become a problem more relevant to ChatGPT, albeit with different characteristics under the same term across various applications.

In recent times, for reasons discussed earlier in this paper as well as broader concern about giving AI "human" characteristics inadvertently by using this term,  researchers have made efforts to replace the term 'hallucination,' deeming it unsuitable and advocating for its renaming or for alternatives. We have compiled many of the suggested terms found in the literature in Table \ref{terms}, along with their definitions in the respective papers. This is a start in the right direction perhaps - in the search for specific definitions and specific characteristics that we want to model - but does illustrates the lack of consistency in the literature that we pointed out in this paper. 

Based on the alternate terms we found, some "old" problems appear to re-surface: the terms confabulation and delusional for instance have connections to mental health conditions as well. However,  fabrication, stochastic parroting and hasty generalization together suggest three viable alternatives. Fact fabrication captures many of the cases previously attributed to 'hallucination' without the negative connotations, while stochastic parroting appears to be an appropriate descriptive term for the reasons behind fact fabrication in Generative AI. While we need clarity in terms of distinguishing between stated facts (in the training data) and inferences, the reference to hasty generalization does start to capture such a distinction.

Finally, since our focus here was on reviewing  AI hallucinations across various applications, we grouped all the final papers examined by category, extracted definitions related to AI hallucination, and used ChatGPT 3.5 \cite{chatgpt3.5} to extract key points. The applications included chatbots, dialogue settings, generative AI, academia, health, legal and ethical settings, science, technology, text translation, question and answering, text summarization, and others. As shown in Table \ref{LLM-Summary}, the extracted summaries share similar characteristics, but highlight different extents of inaccuracy, ranging from "deviating from established knowledge", "factual incorrectness", "fictional" to "nonsensical" -- offering further considerations for a robust taxonomy that will be needed to bring out such nuances.

\begin{table*}[h!]
  \centering
  \scriptsize
  \caption{Key points of "Hallucination" definitions within each application. The characteristics of definitions are presented in \textbf{Bold}, although they may be similar across different applications.}
  \begin{threeparttable} 
         \begin{tabular}{p{5.75em}lccp{55.45em}} 
            \toprule
            \textbf{Application} &       & \multicolumn{1}{p{3em}}{\centering{\textbf{Number of Papers}}} &       & \multicolumn{1}{p{34.115em}}{\textbf{LLM Generated Key Points of Definitions}} \\
            \cmidrule{1-1}\cmidrule{3-3}\cmidrule{5-5}    Chatbot &       & 34    &       & The definitions collectively highlight the central theme of AI-generated content \textbf{deviating from factual correctness}, at times even leading to entirely \textbf{fictional or erroneous information}. In essence, AI hallucination underscores the ongoing challenge of maintaining accuracy and reliability in AI-generated content within the context of chatbot applications. \\
            \cmidrule{1-1}\cmidrule{3-3}\cmidrule{5-5}    Dialogue Setting &       & 8     &       & The definitions collectively underscore the challenge of ensuring accuracy and reliability in dialogue systems, given the potential pitfalls associated with generating content that is \textbf{unsupported, nonsensical, or factually incorrect}. These issues are particularly pertinent when deploying large pre-trained language models in dialogue applications, as they struggle with maintaining fidelity to the source material while generating coherent and accurate responses. \\
            \cmidrule{1-1}\cmidrule{3-3}\cmidrule{5-5}    Generative AI &       & 50    &       & The definitions collectively emphasize the complexity of ensuring factual accuracy and reliability in AI-generated content within generative AI applications, highlighting the potential pitfalls of \textbf{deviating from adherence to factual correctness}. \\
            \cmidrule{1-1}\cmidrule{3-3}\cmidrule{5-5}    Academia &       & 88    &       & A common thread among these definitions is the generation of text or content by AI models that \textbf{lacks fidelity to factual accuracy, reality, or the intended context}. \\
            \cmidrule{1-1}\cmidrule{3-3}\cmidrule{5-5}    Health &       & 82    &       & The key idea common to all the definitions is that "AI hallucination" occurs when AI systems generate information that \textbf{deviates from factual accuracy, context, or established knowledge}. In essence, AI hallucination manifests as the production of text that, though \textbf{potentially plausible, deviates from established facts or knowledge} in health applications. \\
            \cmidrule{1-1}\cmidrule{3-3}\cmidrule{5-5}    Legal and Ethical Setting &       & 16    &       & The definitions collectively emphasize the multifaceted challenges posed by AI hallucination in the legal and ethical context. They highlight \textbf{issues of accuracy, confidence, relevance, context, and potential misinformation}, underscoring the critical importance of addressing these challenges to ensure the responsible and ethical use of AI systems. \\
            \cmidrule{1-1}\cmidrule{3-3}\cmidrule{5-5}    Science &       & 10    &       & Across the definitions, the central theme is that AI hallucination involves the generation of text or information that \textbf{deviates from factual accuracy, coherence, or faithfulness to the input or source content}, with potential consequences for scientific accuracy and integrity. \\
            \cmidrule{1-1}\cmidrule{3-3}\cmidrule{5-5}    Technology &       & 8     &       & The definitions reflect the multifaceted nature of AI hallucination in technology applications, encompassing \textbf{accuracy, unpredictability, credibility, and the balance between reasonableness and correctness}. \\
            \cmidrule{1-1}\cmidrule{3-3}\cmidrule{5-5}    Text Translation &       & 4     &       & The definitions collectively emphasize the central theme of "AI hallucination" in text translation, which revolves around challenges related to maintaining \textbf{fidelity, coherence, and relevance} in the generated translations to ensure accurate and meaningful output. \\
            \cmidrule{1-1}\cmidrule{3-3}\cmidrule{5-5}    Question and Answering &       & 7     &       & "AI hallucination" in question and answer applications raises concerns related to the \textbf{accuracy, truthfulness, and potential spread of misinformation} in AI-generated answers, emphasizing the need for improving the reliability of these systems. \\
            \cmidrule{1-1}\cmidrule{3-3}\cmidrule{5-5}    Text Summarization &       & 19    &       & The definitions highlight the multifaceted challenges posed by "AI hallucination" in text summarization, encompassing issues related to \textbf{fidelity, coherence, factual accuracy, and the preservation of the original meaning} in generated summaries. \\
            \cmidrule{1-1}\cmidrule{3-3}\cmidrule{5-5}    Others \tnote{*} &       & 7     &       & These diverse applications collectively emphasize the challenge of maintaining \textbf{accuracy, coherence, and trustworthiness} in AI-generated content, highlighting the need for tailored approaches to address domain-specific concerns. \\
                \bottomrule
        \end{tabular}%
        \begin{tablenotes}[normal,flushleft]
            \item [*] Including: Investment portfolio, Journalism, Reinforcement Learning, Retail, Sport, and Survey Setting.
        \end{tablenotes}
  \end{threeparttable}
  \label{LLM-Summary}%
\end{table*}%

\section{Discussion}
\begin{table*}[!h]
  \scriptsize
  \centering
  \caption{Some popular press articles on AI hallucination}
  \label{press_article}%
    \begin{tabular}{crp{29.5em}lp{25em}rp{6em}}
\cmidrule{3-7}          &       & \textbf{What the Press Article Discussed…} &       & \textbf{The Real Meaning the Press Article Conveys about "AI Hallucination"} &       & \textbf{Source} \\
\cmidrule{1-1}\cmidrule{3-3}\cmidrule{5-5}\cmidrule{7-7}    1     &       & CNBC provided some examples where ChatGPT generated outputs that sounded correct but weren't actually true, such as a legal brief written by ChatGPT to a Manhattan federal judge &       & When an AI model “hallucinates,” it generates fabricated information in response to a user’s prompt, but presents it as if it’s factual and correct &       & CNBC \cite{CNBC} \\
\cmidrule{1-1}\cmidrule{3-3}\cmidrule{5-5}\cmidrule{7-7}    2     &       & The New York Times asked ChatGPT, Google's Brad, and Microsoft’s Bing: When did The New York Times first report on "artificial intelligence"? &       & Chatbots provide inaccurate answers to questions; although false, the responses appear plausible as they blur and conflate people, events, and ideas &       & The New York Times \cite{TNYT} \\
\cmidrule{1-1}\cmidrule{3-3}\cmidrule{5-5}\cmidrule{7-7}    3     &       & The New York Times traced the evolution of the term "hallucination" throughout the newspaper's history &       & -     &       & The New York Times \cite{TNYT1} \\
\cmidrule{1-1}\cmidrule{3-3}\cmidrule{5-5}\cmidrule{7-7}    4     &       & CNN addressed the major issue of "AI hallucination" and narrated on the responses of OpenAI's and Google's CEOs to the question: Can hallucination be prevented? &       & AI-powered tools like ChatGPT impress with their ability to provide human-like responses, but a growing concern is their tendency to just make things up &       & CNN \cite{CNN} \\
\cmidrule{1-1}\cmidrule{3-3}\cmidrule{5-5}\cmidrule{7-7}    5     &       & Forbes narrated the history of artificial neural networks, which started around eight decades ago, when researchers sought to replicate the functioning of the brain &       & "AI hallucination" refers to unrealistic ideas about achieving "artificial general intelligence" (AGI), while understanding of how our brains work is limited &       & Forbes \cite{Forbes} \\
    \bottomrule
    \end{tabular}%
\end{table*}%

“Hallucinate” secured its position as the word of 2023 (\cite{CambridgeDictionary}, \cite{Dictionary}) and Dictionary.com noted a 46\% surge in searches for the term over the past year. The popular press has also keyed in on this (Table \ref{press_article} highlights topics some recent articles discuss and the meaning of "AI hallucination" they convey). These articles primarily feature interviews with CEOs of big tech companies, who discuss future efforts to prevent "hallucination" in chatbots' outputs. Indeed, preventing such occurrences continues to be a key research goal, but few solutions have emerged so far. In the meanwhile, the Generative AI continues to expand its applications into multiple domains, making the need for good solutions vital. As a precursor to even developing solutions, this paper calls for more systematic, consistent and semantically nuanced terms that can replace "hallucinations" for the reasons noted here. As one step toward such a call, we presented a short summary from one of the broadest manual literature reviews on this topic to date. Our findings illustrate the current lack of consistency and consensus on this issue, but also bring to light some recent options that are good alternatives. More work is needed to develop a systematic taxonomy that can be widely adopted as we discuss these issues in the context of AI applications.


\appendix[Review of the "AI Hallucination" definitions]
\onecolumn
\footnotesize
\begin{longtable}{p{2em}p{0.05em}p{7em}p{0.05em}p{2em}p{0.05em}p{5em}p{0.05em}p{21.95em}p{0.05em}p{2em}}
\caption{AI hallucination definitions} \label{review_table} \\

\toprule
\textbf{Num.} &       & \multicolumn{1}{c}{\textbf{Author(s)}} &       & \textbf{Year} &       & \multicolumn{1}{c}{\textbf{Application}} &       & \multicolumn{1}{c}{\textbf{Definition}}&       & \multicolumn{1}{c}{\textbf{Citation}} \\ 
\endfirsthead

\multicolumn{11}{c}%
{{\bfseries \tablename\ \thetable{} -- continued from previous page}} \\
\toprule
\textbf{Num.} &       & \multicolumn{1}{c}{\textbf{Author(s)}} &       & \textbf{Year} &       & \multicolumn{1}{c}{\textbf{Application}} &       & \multicolumn{1}{c}{\textbf{Definition}}&       & \multicolumn{1}{c}{\textbf{Citation}} \\
\cmidrule{1-1}\cmidrule{3-3}\cmidrule{5-5}\cmidrule{7-7}\cmidrule{9-9}\cmidrule{11-11}
\endhead

\hline \multicolumn{11}{|r|}{{Continued on next page}} \\ \hline
\endfoot

\hline \hline
\endlastfoot

\cmidrule{1-1}\cmidrule{3-3}\cmidrule{5-5}\cmidrule{7-7}\cmidrule{9-9}\cmidrule{11-11}    1     &       & Koehn and Knowles \cite{koehn2017challenges} &       & 2017  &       & Text Translation &       & "AI hallucination" occurs when the output of the Neural Machine Translation (NMT) system is often quite fluent but entirely unrelated to the input. &       & 1249 \\
\cmidrule{1-1}\cmidrule{3-3}\cmidrule{5-5}\cmidrule{7-7}\cmidrule{9-9}\cmidrule{11-11}    2     &       & Wiseman et al. \cite{wiseman2017challenges} &       & 2017  &       & Natural Language Generation &       & "AI hallucination" occurs when a language model presumes likelihood, but the generated content is ultimately incorrect and unsupported by any information. &       & 556 \\
\cmidrule{1-1}\cmidrule{3-3}\cmidrule{5-5}\cmidrule{7-7}\cmidrule{9-9}\cmidrule{11-11}    3     &       & Lee et al. \cite{lee2018hallucinations} &       & 2018  &       & Text Translation &       & "AI hallucination" occurs when Neural Machine Translation (NMT) systems produce highly pathological translations that are completely untethered from the source material. &       & 83 \\
\cmidrule{1-1}\cmidrule{3-3}\cmidrule{5-5}\cmidrule{7-7}\cmidrule{9-9}\cmidrule{11-11}    4     &       & Nie et al. \cite{nie2019simple} &       & 2019  &       & Large Language Model &       & "AI hallucination" refers to the problem where the generated texts often contain information that is irrelevant to or contradicted by the input. &       & 63 \\
\cmidrule{1-1}\cmidrule{3-3}\cmidrule{5-5}\cmidrule{7-7}\cmidrule{9-9}\cmidrule{11-11}    5     &       & Tian et al. \cite{tian2019sticking} &       & 2019  &       & Large Language Model &       & "AI hallucination" occurs when a language model generates text that is fluent but unfaithful to the source. &       & 66 \\
\cmidrule{1-1}\cmidrule{3-3}\cmidrule{5-5}\cmidrule{7-7}\cmidrule{9-9}\cmidrule{11-11}    6     &       & Du{\v{s}}ek et al. \cite{duvsek2019semantic} &       & 2019  &       & Natural Language Generation &       & "AI hallucination" occurs when a language model adds information that is not grounded in the input. &       & 89 \\
\cmidrule{1-1}\cmidrule{3-3}\cmidrule{5-5}\cmidrule{7-7}\cmidrule{9-9}\cmidrule{11-11}    7     &       & Ferreira et al. \cite{ferreira2019neural} &       & 2019  &       & Natural Language Generation &       & "AI hallucination" refers to a language model describing non-linguistic representations that are not present in the input. &       & 141 \\
\cmidrule{1-1}\cmidrule{3-3}\cmidrule{5-5}\cmidrule{7-7}\cmidrule{9-9}\cmidrule{11-11}    8     &       & Martindale et al. \cite{martindale2019identifying} &       & 2019  &       & Text Translation &       & "AI hallucination" occurs when the machine translation output contains more information than the reference text. &       & 29 \\
\cmidrule{1-1}\cmidrule{3-3}\cmidrule{5-5}\cmidrule{7-7}\cmidrule{9-9}\cmidrule{11-11}    9     &       & Du{\v{s}}ek and Kasner \cite{duvsek2020evaluating} &       & 2020  &       & Natural Language Generation &       & "AI hallucination" occurs when expressions in the output do not correspond to input facts. &       & 46 \\
\cmidrule{1-1}\cmidrule{3-3}\cmidrule{5-5}\cmidrule{7-7}\cmidrule{9-9}\cmidrule{11-11}    10    &       & Kang and Hashimoto \cite{kang2020improved} &       & 2020  &       & Natural Language Generation &       & "AI hallucination" occurs when neural language models often produce fluent text that is unfaithful to the source. &       & 72 \\
\cmidrule{1-1}\cmidrule{3-3}\cmidrule{5-5}\cmidrule{7-7}\cmidrule{9-9}\cmidrule{11-11}    11    &       & Parikh et al. \cite{parikh2020totto} &       & 2020  &       & Natural Language Generation &       & "AI hallucination" occurs when a language model generates text that is fluent but not faithful to the source. &       & 242 \\
\cmidrule{1-1}\cmidrule{3-3}\cmidrule{5-5}\cmidrule{7-7}\cmidrule{9-9}\cmidrule{11-11}    12    &       & Maynez et al. \cite{maynez2020faithfulness} &       & 2020  &       & Text Summarization &       & "AI hallucination" refers to content that is unfaithful to the input document.\newline{}"Intrinsic hallucinations" are consequences of synthesizing content using the information present in the input document.\newline{}"Extrinsic hallucinations" are model generations that ignore the source material altogether. Hallucinate content that is unfaithful to the input document. &       & 572 \\
\cmidrule{1-1}\cmidrule{3-3}\cmidrule{5-5}\cmidrule{7-7}\cmidrule{9-9}\cmidrule{11-11}    13    &       & Durmus et al. \cite{durmus2020feqa} &       & 2020  &       & Question Answering &       & "AI hallucination" refers to the occurrence when AI generates content inconsistent with the source document, i.e., unfaithful. &       & 269 \\
\cmidrule{1-1}\cmidrule{3-3}\cmidrule{5-5}\cmidrule{7-7}\cmidrule{9-9}\cmidrule{11-11}    14    &       & Zhao et al. \cite{zhao2020reducing} &       & 2020  &       & Text Summarization &       & "AI hallucination" occurs when language models generate material that is not supported by the original text. &       & 73 \\
\cmidrule{1-1}\cmidrule{3-3}\cmidrule{5-5}\cmidrule{7-7}\cmidrule{9-9}\cmidrule{11-11}    15    &       & Dong et al. \cite{dong2020multi} &       & 2020  &       & Text Summarization &       & "AI hallucination" occurs when a language model generates content that is factually inconsistent with the source documents. &       & 89 \\
\cmidrule{1-1}\cmidrule{3-3}\cmidrule{5-5}\cmidrule{7-7}\cmidrule{9-9}\cmidrule{11-11}    16    &       & Filippova \cite{filippova2020controlled} &       & 2020  &       & Large Language Model &       & "AI hallucination" refers to the generated content which is either unfaithful to the input or nonsensical. &       & 46 \\
\cmidrule{1-1}\cmidrule{3-3}\cmidrule{5-5}\cmidrule{7-7}\cmidrule{9-9}\cmidrule{11-11}    17    &       & Zhou et al. \cite{zhou2020detecting} &       & 2020  &       & Hallucination Detection &       & "AI hallucination" occurs when the model generates additional content not supported by the input. &       & 87 \\
\cmidrule{1-1}\cmidrule{3-3}\cmidrule{5-5}\cmidrule{7-7}\cmidrule{9-9}\cmidrule{11-11}    18    &       & Elsahar et al. \cite{elsahar2020self} &       & 2020  &       & Hallucination Mitigation &       & "AI hallucination" occurs when the fluency of natural language generation models can be highly misleading, as it often distracts from the wrong facts stated in the generated text. &       & 33 \\
\cmidrule{1-1}\cmidrule{3-3}\cmidrule{5-5}\cmidrule{7-7}\cmidrule{9-9}\cmidrule{11-11}    19    &       & Tang et al. \cite{tang2021confit} &       & 2021  &       & Text Summarization &       & "AI hallucination" refers to content that is not supported by the source documents. &       & 32 \\
\cmidrule{1-1}\cmidrule{3-3}\cmidrule{5-5}\cmidrule{7-7}\cmidrule{9-9}\cmidrule{11-11}    20    &       & Cao et al. \cite{cao2021hallucinated} &       & 2021  &       & Text Summarization &       & "AI hallucination" refers to content that is not directly inferable from the source text. &       & 37 \\
\cmidrule{1-1}\cmidrule{3-3}\cmidrule{5-5}\cmidrule{7-7}\cmidrule{9-9}\cmidrule{11-11}    21    &       & Miao et al. \cite{miao2021prevent} &       & 2021  &       & Text Translation &       & "AI hallucination" refers to generating fluent but inadequate translations to the source sentences. &       & 21 \\
\cmidrule{1-1}\cmidrule{3-3}\cmidrule{5-5}\cmidrule{7-7}\cmidrule{9-9}\cmidrule{11-11}    22    &       & Chen et al. \cite{chen2021improving} &       & 2021  &       & Text Summarization &       & "Intrinsic" and "Extrinsic hallucinations", involving the fabrication of untruthful text spans containing information that may either be present or absent from the source. &       & 62 \\
\cmidrule{1-1}\cmidrule{3-3}\cmidrule{5-5}\cmidrule{7-7}\cmidrule{9-9}\cmidrule{11-11}    23    &       & Wang et al. \cite{wang2021exploring} &       & 2021  &       & Text Summarization &       & "AI hallucination" refers to where the model generates fictional content. &       & 14 \\
\cmidrule{1-1}\cmidrule{3-3}\cmidrule{5-5}\cmidrule{7-7}\cmidrule{9-9}\cmidrule{11-11}    24    &       & Xiao and Wang \cite{xiao2021hallucination} &       & 2021  &       & Large Language Model &       & "AI hallucination" occurs where models generate description tokens that are not supported by the source inputs. &       & 58 \\
\cmidrule{1-1}\cmidrule{3-3}\cmidrule{5-5}\cmidrule{7-7}\cmidrule{9-9}\cmidrule{11-11}    25    &       & Dziri et al. \cite{dziri2021neural} &       & 2021  &       & Knowledge Graph &       & "AI hallucination" occurs when dialogue models, despite maintaining plausible general linguistic capabilities, are still unable to fully discern facts and may instead hallucinate factually invalid information. &       & 51 \\
\cmidrule{1-1}\cmidrule{3-3}\cmidrule{5-5}\cmidrule{7-7}\cmidrule{9-9}\cmidrule{11-11}    26    &       & Liu et al. \cite{liu2021token} &       & 2021  &       & Hallucination Detection &       & "AI hallucination" occurs when language models exhibit a propensity to hallucinate non-existent or incorrect content that is unacceptable in most user-oriented applications. &       & 29 \\
\cmidrule{1-1}\cmidrule{3-3}\cmidrule{5-5}\cmidrule{7-7}\cmidrule{9-9}\cmidrule{11-11}    27    &       & Huang et al. \cite{huang2021factual} &       & 2021  &       & Text Summarization &       & "Intrinsic hallucination" is a fact that is contradicted to the source document.\newline{}"Extrinsic hallucination" is the fact that is neutral to the source document (i.e., the content that is neither supported nor contradicted by the source document). &       & 51 \\
\cmidrule{1-1}\cmidrule{3-3}\cmidrule{5-5}\cmidrule{7-7}\cmidrule{9-9}\cmidrule{11-11}    28    &       & Lin \cite{lin2021truthfulqa} &       & 2021  &       & Question Answering &       & "AI hallucination" refers to the tendency of LLMs to generate false statements. &       & 250 \\
\cmidrule{1-1}\cmidrule{3-3}\cmidrule{5-5}\cmidrule{7-7}\cmidrule{9-9}\cmidrule{11-11}    29    &       & Shuster et al. \cite{shuster2021retrieval} &       & 2021  &       & Large Language Model &       & "AI hallucination" refers to the occurrence where language models generate plausible-looking statements that are factually incorrect. &       & 198 \\
\cmidrule{1-1}\cmidrule{3-3}\cmidrule{5-5}\cmidrule{7-7}\cmidrule{9-9}\cmidrule{11-11}    30    &       & Sekuli{\'c} et al. \cite{sekulic2021towards} &       & 2021  &       & Natural Language Generation &       & "AI hallucination" occurs where generated responses do not correspond to the real-world. &       & 26 \\
\cmidrule{1-1}\cmidrule{3-3}\cmidrule{5-5}\cmidrule{7-7}\cmidrule{9-9}\cmidrule{11-11}    31    &       & Ghosh et al. \cite{ghosh2021helpful} &       & 2021  &       & Reinforcement Learning &       & "AI hallucination" occurs when the generated text asserts information not present in the source. &       & 5 \\
\cmidrule{1-1}\cmidrule{3-3}\cmidrule{5-5}\cmidrule{7-7}\cmidrule{9-9}\cmidrule{11-11}    32    &       & Perez-Beltrachini and Lapata \cite{perez2021multi} &       & 2021  &       & Text Summarization &       & "AI hallucination" refers to neural summarization models' propensity to generate text that does not preserve the meaning of the input. &       & 18 \\
\cmidrule{1-1}\cmidrule{3-3}\cmidrule{5-5}\cmidrule{7-7}\cmidrule{9-9}\cmidrule{11-11}    33    &       & Lyu et al. \cite{lyu2022faithful} &       & 2022  &       & Text Summarization &       & "AI hallucination" signifies the presence of distorting or fabricating facts within generated summaries, resulting in inconsistencies between a summary and the corresponding original document.\newline{}"Extrinsic hallucination" entails adding information not directly inferable from the input information.\newline{}"Intrinsic hallucination" involves manipulating the information present in the input document. &       & 0 \\
\cmidrule{1-1}\cmidrule{3-3}\cmidrule{5-5}\cmidrule{7-7}\cmidrule{9-9}\cmidrule{11-11}    34    &       & Ali et al. \cite{ali2022performance} &       & 2022  &       & Health &       & "AI hallucination" refers to scenarios in which a Language Model (LLM) asserts inaccurate facts or contextual data that it falsely believes to be correct in its response. &       & 29 \\
\cmidrule{1-1}\cmidrule{3-3}\cmidrule{5-5}\cmidrule{7-7}\cmidrule{9-9}\cmidrule{11-11}    35    &       & Ando et al. \cite{ando2022artificial} &       & 2022  &       & Health &       & "AI hallucination" occurs when abstractive summarization may sometimes unintentionally generate unfaithful descriptions.\newline{}"Intrinsic hallucination" is a phenomenon in which the concept or term itself is in the source documents; its synthesis misrepresents the information in the source, and the meaning becomes inconsistent.\newline{}"Extrinsic hallucination" is content that is neither supported nor contradicted by the source and is caused by source documents with poor information. &       & 2 \\
\cmidrule{1-1}\cmidrule{3-3}\cmidrule{5-5}\cmidrule{7-7}\cmidrule{9-9}\cmidrule{11-11}    36    &       & Ando et al. \cite{ando2022exploring} &       & 2022  &       & Health &       & "AI hallucination" refers to the phenomenon where the abstractive approach frequently produces fake content that does not align with the reference summary. &       & 2 \\
\cmidrule{1-1}\cmidrule{3-3}\cmidrule{5-5}\cmidrule{7-7}\cmidrule{9-9}\cmidrule{11-11}    37    &       & Rebuffel et al. \cite{rebuffel2022controlling} &       & 2022  &       & Natural Language Generation &       & "AI hallucination" refers to model outputs that are often subject to over-generation, where misaligned fragments from training instances, known as divergences, can induce similarly misaligned outputs during inference. &       & 43 \\
\cmidrule{1-1}\cmidrule{3-3}\cmidrule{5-5}\cmidrule{7-7}\cmidrule{9-9}\cmidrule{11-11}    38    &       & Wan and Bansal \cite{wan2022factpegasus} &       & 2022  &       & Text Summarization &       & "AI hallucination" refers to a scenario where a summary contains facts or entities not present in the original document. &       & 30 \\
\cmidrule{1-1}\cmidrule{3-3}\cmidrule{5-5}\cmidrule{7-7}\cmidrule{9-9}\cmidrule{11-11}    39    &       & Corbelle et al. \cite{corbelle2022dealing} &       & 2022  &       & Natural Language Generation &       & "AI hallucination" refers to instances where neural models generate texts that are incoherent or unrelated to the input of a D2T system. &       & 2 \\
\cmidrule{1-1}\cmidrule{3-3}\cmidrule{5-5}\cmidrule{7-7}\cmidrule{9-9}\cmidrule{11-11}    40    &       & Lee et al. \cite{lee2022factuality} &       & 2022  &       & Large Language Model &       & "AI hallucination" occurs when a model is making factual errors, generating a named-entity that does not appear in the ground-truth knowledge source. &       & 30 \\
\cmidrule{1-1}\cmidrule{3-3}\cmidrule{5-5}\cmidrule{7-7}\cmidrule{9-9}\cmidrule{11-11}    41    &       & Cabezudo et al. \cite{cabezudo2022exploring} &       & 2022  &       & Natural Language Generation &       & "AI hallucination" refers to text generated by pre-trained models that is irrelevant to or contradicted with the input. &       & 0 \\
\cmidrule{1-1}\cmidrule{3-3}\cmidrule{5-5}\cmidrule{7-7}\cmidrule{9-9}\cmidrule{11-11}    42    &       & Van et al. \cite{van2022mutual} &       & 2022  &       & Text Summarization &       & "AI hallucination" refers to the occurrence of adding information to the output that was not present in the original text. &       & 7 \\
\cmidrule{1-1}\cmidrule{3-3}\cmidrule{5-5}\cmidrule{7-7}\cmidrule{9-9}\cmidrule{11-11}    43    &       & Dziri et al. \cite{dziri2022origin} &       & 2022  &       & Dialogue &       & "AI hallucination" occurs when large pre-trained language models generate factually incorrect statements. &       & 51 \\
\cmidrule{1-1}\cmidrule{3-3}\cmidrule{5-5}\cmidrule{7-7}\cmidrule{9-9}\cmidrule{11-11}    44    &       & Dziri et al. \cite{dziri2022faithdial} &       & 2022  &       & Dialogue &       & "AI hallucination" refers to the phenomenon where dialogue systems often produce unsupported utterances. &       & 24 \\
\cmidrule{1-1}\cmidrule{3-3}\cmidrule{5-5}\cmidrule{7-7}\cmidrule{9-9}\cmidrule{11-11}    45    &       & Koto et al. \cite{koto2022ffci} &       & 2022  &       & Text Summarization &       & "AI hallucination" occurs where information is generated that does not exist in the source document, also called "factual inconsistencies." &       & 21 \\
\cmidrule{1-1}\cmidrule{3-3}\cmidrule{5-5}\cmidrule{7-7}\cmidrule{9-9}\cmidrule{11-11}    46    &       & Goodman et al. \cite{goodman2022lampost} &       & 2022  &       & Email Writing &       & "AI hallucination" refers to factually incorrect or non-existent content generated by the LLM. &       & 10 \\
\cmidrule{1-1}\cmidrule{3-3}\cmidrule{5-5}\cmidrule{7-7}\cmidrule{9-9}\cmidrule{11-11}    47    &       & Gehrmann et al. \cite{gehrmann2023repairing} &       & 2022  &       & Natural Language Generation &       & "AI hallucination" refers to a situation where a model is not faithful as it adds information not present in the source document.\newline{}"Intrinsic hallucinations" misrepresent facts in the input. \newline{}"Extrinsic hallucinations" ignore the input altogether. &       & 62 \\
\cmidrule{1-1}\cmidrule{3-3}\cmidrule{5-5}\cmidrule{7-7}\cmidrule{9-9}\cmidrule{11-11}    48    &       & Erdem et al. \cite{erdem2022neural} &       & 2022  &       & Natural Language Generation &       & "AI hallucination" refers to the generation of descriptions or facts that are not fully supported by the input. &       & 20 \\
\cmidrule{1-1}\cmidrule{3-3}\cmidrule{5-5}\cmidrule{7-7}\cmidrule{9-9}\cmidrule{11-11}    49    &       & Tun et al. \cite{tun2022comex} &       & 2022  &       & Dialogue &       & "AI hallucination" refers to large-scale pre-trained language models generating text that is nonsensical and struggling to remain true to the source content. &       & 0 \\
\cmidrule{1-1}\cmidrule{3-3}\cmidrule{5-5}\cmidrule{7-7}\cmidrule{9-9}\cmidrule{11-11}    50    &       & Gurrapu et al. \cite{gurrapu2022exclaim} &       & 2022  &       & Claim Verification &       & "AI hallucination" refers to the phenomenon where natural language generation models introduce unintended and irrelevant text during the generation process. &       & 1 \\
\cmidrule{1-1}\cmidrule{3-3}\cmidrule{5-5}\cmidrule{7-7}\cmidrule{9-9}\cmidrule{11-11}    51    &       & Yang et al. \cite{yang2022explainable} &       & 2022  &       & Text Summarization &       & "AI hallucination" occurs when models generate summaries factually inconsistent with their original document. &       & 2 \\
\cmidrule{1-1}\cmidrule{3-3}\cmidrule{5-5}\cmidrule{7-7}\cmidrule{9-9}\cmidrule{11-11}    52    &       & Ji et al. \cite{ji2023survey} &       & 2023  &       & Natural Language Generation &       & "AI hallucination" refers to the generated content that is nonsensical or unfaithful to the provided source content.\newline{}"Intrinsic hallucinations" encompasses the generated output that contradicts the source content.\newline{}"Extrinsic hallucination" involves the generated output that cannot be verified from the source content (i.e., output can neither be supported nor contradicted by the source). &       & 482 \\
\cmidrule{1-1}\cmidrule{3-3}\cmidrule{5-5}\cmidrule{7-7}\cmidrule{9-9}\cmidrule{11-11}    53    &       & Narayanan et al. \cite{narayanan2023unmasking} &       & 2023  &       & Natural Language Generation &       & "AI hallucination" refers to situations in which AI models provide responses with confidence that appear faithful but are nonsensical when evaluated against common knowledge. &       & 2 \\
\cmidrule{1-1}\cmidrule{3-3}\cmidrule{5-5}\cmidrule{7-7}\cmidrule{9-9}\cmidrule{11-11}    54    &       & Leiser et al. \cite{leiser2023chatgpt} &       & 2023  &       & Large Language Model &       & "Artificial hallucinations" are the distortion of perception in Large Language Models (LLMs) as they present incorrect information, disguising it as factual responses. &       & 0 \\
\cmidrule{1-1}\cmidrule{3-3}\cmidrule{5-5}\cmidrule{7-7}\cmidrule{9-9}\cmidrule{11-11}    55    &       & Wang et al. \cite{wang2023human} &       & 2023  &       & ChatGPT &       & "AI hallucination" denotes the tendency to generate inaccurate outputs unfaithful to the training data. &       & 1 \\
\cmidrule{1-1}\cmidrule{3-3}\cmidrule{5-5}\cmidrule{7-7}\cmidrule{9-9}\cmidrule{11-11}    56    &       & Ma et al. \cite{ma2023scope} &       & 2023  &       & Software Engineering &       & "AI hallucination" refers to the phenomenon that can result in the fabrication of elements that do not actually exist. &       & 8 \\
\cmidrule{1-1}\cmidrule{3-3}\cmidrule{5-5}\cmidrule{7-7}\cmidrule{9-9}\cmidrule{11-11}    57    &       & Li \cite{li2023dark} &       & 2023  &       & Large Language Model &       & "AI hallucination" occurs when LLMs generate text based on their internal logic or patterns, rather than the true context, leading to confidently but unjustified and unverified deceptive responses. &       & 11 \\
\cmidrule{1-1}\cmidrule{3-3}\cmidrule{5-5}\cmidrule{7-7}\cmidrule{9-9}\cmidrule{11-11}    58    &       & Vaghefi et al. \cite{vaghefi2023chatclimate} &       & 2023  &       & NLP and Climate Change &       & "AI hallucination" refers to mistakes in the generated text that are semantically incorrect or unsupported by the input text. &       & 5 \\
\cmidrule{1-1}\cmidrule{3-3}\cmidrule{5-5}\cmidrule{7-7}\cmidrule{9-9}\cmidrule{11-11}    59    &       & Daull et al. \cite{daull2023complex} &       & 2023  &       & Question Answering &       & "AI hallucination" refers to confident generated responses that contain false information not supported by the model’s training data.\newline{}"Extrinsic hallucination" involves a model introducing information not present in the source data.\newline{}"Intrinsic hallucination" occurs when the model distorts information from the source data into a factually incorrect representation. &       & 2 \\
\cmidrule{1-1}\cmidrule{3-3}\cmidrule{5-5}\cmidrule{7-7}\cmidrule{9-9}\cmidrule{11-11}    60    &       & Zhang et al. \cite{zhang2023siren} &       & 2023  &       & Large Language Model &       & "AI hallucination" commonly refers to the phenomenon where LLMs occasionally generate outputs that, although appearing plausible, deviate from the user input, previously generated context, or factual knowledge. &       & 3 \\
\cmidrule{1-1}\cmidrule{3-3}\cmidrule{5-5}\cmidrule{7-7}\cmidrule{9-9}\cmidrule{11-11}    61    &       & Romanko et al. \cite{romanko2023chatgpt} &       & 2023  &       & Investment Portfolio &       & "AI hallucination" refers to instances where the AI, although generating text based on its training, does so without a solid conceptual framework behind it. &       & 3 \\
\cmidrule{1-1}\cmidrule{3-3}\cmidrule{5-5}\cmidrule{7-7}\cmidrule{9-9}\cmidrule{11-11}    62    &       & Henderson et al. \cite{henderson2023s} &       & 2023  &       & AI Speech &       & "AI hallucination" refers to the act of generating text that includes factual claims that are untrue, and in some cases, these claims may not have been present in its training data. &       & 2 \\
\cmidrule{1-1}\cmidrule{3-3}\cmidrule{5-5}\cmidrule{7-7}\cmidrule{9-9}\cmidrule{11-11}    63    &       & Ni et al. \cite{ni2023chatreport} &       & 2023  &       & NLP and Climate Change &       &  "AI hallucination" refers to the generation of answers in which some or all of the covered information is not adequately supported by the report, particularly when there is extrapolation or partial support involved, and it may also relate to instances where the model fails to honestly report its references while generating content that lacks fidelity to the source. &       & 0 \\
\cmidrule{1-1}\cmidrule{3-3}\cmidrule{5-5}\cmidrule{7-7}\cmidrule{9-9}\cmidrule{11-11}    64    &       & Mahmood et al. \cite{mahmood2023fact} &       & 2023  &       & Radiology Report &       & "AI hallucination" involves the generation of false findings in the produced reports by LLMs. &       & 1 \\
\cmidrule{1-1}\cmidrule{3-3}\cmidrule{5-5}\cmidrule{7-7}\cmidrule{9-9}\cmidrule{11-11}    65    &       & Goyal et al. \cite{goyal2023chatgpt} &       & 2023  &       & Question Answering &       & "Hallucinations" signify the production of convincing yet incorrect text outputs by LLMs, with the potential to distort scientific facts and disseminate misinformation. &       & 0 \\
\cmidrule{1-1}\cmidrule{3-3}\cmidrule{5-5}\cmidrule{7-7}\cmidrule{9-9}\cmidrule{11-11}    66    &       & Zhang \cite{zhang2023bridging} &       & 2023  &       & Robotic &       & "AI hallucination" refers to unpredictable outputs generated by LLMs. &       & 0 \\
\cmidrule{1-1}\cmidrule{3-3}\cmidrule{5-5}\cmidrule{7-7}\cmidrule{9-9}\cmidrule{11-11}    67    &       & Li et al. \cite{li2023trac} &       & 2023  &       & Large Language Model &       & "AI hallucination" is the phenomenon where responses are confidently generated but are incorrect. &       & 1 \\
\cmidrule{1-1}\cmidrule{3-3}\cmidrule{5-5}\cmidrule{7-7}\cmidrule{9-9}\cmidrule{11-11}    68    &       & Li et al. \cite{li2023batgpt} &       & 2023  &       & Large Language Model &       & "AI hallucination" occurs when AI systems generate outputs that are not aligned with the input context and encounter challenges in efficiently capturing complex dependencies. &       & 1 \\
\cmidrule{1-1}\cmidrule{3-3}\cmidrule{5-5}\cmidrule{7-7}\cmidrule{9-9}\cmidrule{11-11}    69    &       & Curran et al. \cite{curran2023hallucination} &       & 2023  &       & Legal Setting &       & "AI hallucination" refers to tendency for generating false information. &       & 1 \\
\cmidrule{1-1}\cmidrule{3-3}\cmidrule{5-5}\cmidrule{7-7}\cmidrule{9-9}\cmidrule{11-11}    70    &       & Feldman et al. \cite{feldman2023trapping} &       & 2023  &       & Large Language Model &       & "AI hallucination" in the context of LLMs refers to the phenomenon where these language models generate non-factual statements, which could have a detrimental effect on the trustworthiness of their outputs. &       & 2 \\
\cmidrule{1-1}\cmidrule{3-3}\cmidrule{5-5}\cmidrule{7-7}\cmidrule{9-9}\cmidrule{11-11}    71    &       & Mukherjee et al. \cite{mukherjee2023creative} &       & 2023  &       & Generative AI &       & "Hallucinations," characterized by AI responses containing random inaccuracies or falsehoods, emerge when models prioritize novelty over usefulness. &       & 0 \\
\cmidrule{1-1}\cmidrule{3-3}\cmidrule{5-5}\cmidrule{7-7}\cmidrule{9-9}\cmidrule{11-11}    72    &       & Salvagno et al. \cite{salvagno2023artificial} &       & 2023  &       & Large Language Model &       & "AI hallucination" refers to the creation of seemingly accurate bibliographic references with recognized authors and coherent titles, even though these references are entirely fictitious and have no real existence. &       & 4 \\
\cmidrule{1-1}\cmidrule{3-3}\cmidrule{5-5}\cmidrule{7-7}\cmidrule{9-9}\cmidrule{11-11}    73    &       & Beutel et al. \cite{beutel2023artificial} &       & 2023  &       & Large Language Model &       & "Hallucinations" are defined as the production of content that does not accurately represent the provided source and may appear nonsensical due to errors in the encoding and decoding processes between text and representations. &       & 20 \\
\cmidrule{1-1}\cmidrule{3-3}\cmidrule{5-5}\cmidrule{7-7}\cmidrule{9-9}\cmidrule{11-11}    74    &       & Azamfirei et al. \cite{azamfirei2023large} &       & 2023  &       & Text Summarization &       & "AI hallucination" refers to receiving a response from the model when it lacks an appropriate answer, which appears to be the most likely summary of the study, despite its potential inaccuracy. &       & 32 \\
\cmidrule{1-1}\cmidrule{3-3}\cmidrule{5-5}\cmidrule{7-7}\cmidrule{9-9}\cmidrule{11-11}    75    &       & Meyer et al. \cite{meyer2023chatgpt} &       & 2023  &       & Academia &       & "AI hallucination" can be defined as the capability of LLM-based chatbots to convey false information as though it were true. &       & 8 \\
\cmidrule{1-1}\cmidrule{3-3}\cmidrule{5-5}\cmidrule{7-7}\cmidrule{9-9}\cmidrule{11-11}    76    &       & Hernigou and Scarlat \cite{hernigou2023two} &       & 2023  &       & Health &       & "AI hallucination" is a confident response that does not seem in concordance with its training data. &       & 2 \\
\cmidrule{1-1}\cmidrule{3-3}\cmidrule{5-5}\cmidrule{7-7}\cmidrule{9-9}\cmidrule{11-11}    77    &       & Patil \cite{EMERGINGTECHNOLOGY} &       & 2023  &       & Legal Setting &       & "AI hallucination" is defined as a confident response by an AI system that lacks justification in its training data. These responses can appear factual but are not true, often simply being answers "made up" by the AI. &       & 0 \\
\cmidrule{1-1}\cmidrule{3-3}\cmidrule{5-5}\cmidrule{7-7}\cmidrule{9-9}\cmidrule{11-11}    78    &       & Alexander et al. \cite{alexander2023wrote} &       & 2023  &       & Academia &       & "AI hallucination" refers to its tendency to make up facts and references that do not exist. &       & 1 \\
\cmidrule{1-1}\cmidrule{3-3}\cmidrule{5-5}\cmidrule{7-7}\cmidrule{9-9}\cmidrule{11-11}    79    &       & Lyell \cite{lyell2023history} &       & 2023  &       & Academia &       & "AI hallucination" describes the propensity of the system to convincingly fabricate information. &       & 0 \\
\cmidrule{1-1}\cmidrule{3-3}\cmidrule{5-5}\cmidrule{7-7}\cmidrule{9-9}\cmidrule{11-11}    80    &       & Grassini \cite{grassini2023shaping} &       & 2023  &       & Academia &       & "AI hallucination" involves the generation of incorrect or even fabricated information. &       & 14 \\
\cmidrule{1-1}\cmidrule{3-3}\cmidrule{5-5}\cmidrule{7-7}\cmidrule{9-9}\cmidrule{11-11}    81    &       & Brodeur et al. \cite{brodeur2023chatgpt} &       & 2023  &       & Legal Setting &       & "AI hallucination" refers to the situation where an AI system, due to its inability to correctly interpret data, generates inaccurate or unusual outputs. &       & 0 \\
\cmidrule{1-1}\cmidrule{3-3}\cmidrule{5-5}\cmidrule{7-7}\cmidrule{9-9}\cmidrule{11-11}    82    &       & Lee and Choi \cite{lee2023utilizing} &       & 2023  &       & Health &       & "AI hallucination" occurs when AI generates information about things that are untrue, presenting them as if they are true, which can diminish the reliability of the generated information. &       & 0 \\
\cmidrule{1-1}\cmidrule{3-3}\cmidrule{5-5}\cmidrule{7-7}\cmidrule{9-9}\cmidrule{11-11}    83    &       & Chatfield \cite{newphilosopher2023} &       & 2023  &       & Chatbot &       & "AI hallucination" represents a response that may or may not be plausible but is not rooted in reality. &       & 0 \\
\cmidrule{1-1}\cmidrule{3-3}\cmidrule{5-5}\cmidrule{7-7}\cmidrule{9-9}\cmidrule{11-11}    84    &       & McGowan et al. \cite{mcgowan2023chatgpt} &       & 2023  &       & References &       & "AI hallucination" refers to the occurrence of mistakes in the generated text that, while semantically or syntactically plausible, are ultimately incorrect or nonsensical upon closer examination. &       & 4 \\
\cmidrule{1-1}\cmidrule{3-3}\cmidrule{5-5}\cmidrule{7-7}\cmidrule{9-9}\cmidrule{11-11}    85    &       & Wu and Dang \cite{wu2023chatgpt} &       & 2023  &       & Academia &       & "AI hallucination" refers to a scenario in which a machine generates seemingly realistic outputs without any real-world input. &       & 6 \\
\cmidrule{1-1}\cmidrule{3-3}\cmidrule{5-5}\cmidrule{7-7}\cmidrule{9-9}\cmidrule{11-11}    86    &       & Wang et al. \cite{wang2023performance} &       & 2023  &       & Academia &       & "AI hallucination" refers to the generation of content that is nonsensical or untrue in relation to certain sources. &       & 2 \\
\cmidrule{1-1}\cmidrule{3-3}\cmidrule{5-5}\cmidrule{7-7}\cmidrule{9-9}\cmidrule{11-11}    87    &       & Hou and Ji \cite{hou2023geneturing} &       & 2023  &       & Health &       & "AI hallucination" refers to the situation where GPT models provide confident answers that contradict the truth. &       & 3 \\
\cmidrule{1-1}\cmidrule{3-3}\cmidrule{5-5}\cmidrule{7-7}\cmidrule{9-9}\cmidrule{11-11}    88    &       & Au et al. \cite{au2023ai} &       & 2023  &       & Health &       & "AI hallucination" refers to the phenomenon in which LLMs produce output that is nonsensical or unfaithful to the provided input or "prompt," and this problem is further amplified when the prompt contains insufficient or masked information, despite the high confidence displayed in the generated output, as observed in ChatGPT. &       & 15 \\
\cmidrule{1-1}\cmidrule{3-3}\cmidrule{5-5}\cmidrule{7-7}\cmidrule{9-9}\cmidrule{11-11}    89    &       & Hua et al. \cite{hua2023evaluation} &       & 2023  &       & Health/ Academia &       & "AI hallucination" is AI-generated outputs that deviate from its training data. These outputs may appear syntactically or semantically plausible, but in reality, they are incorrect or nonsensical. &       & 1 \\
\cmidrule{1-1}\cmidrule{3-3}\cmidrule{5-5}\cmidrule{7-7}\cmidrule{9-9}\cmidrule{11-11}    90    &       & Brameier et al. \cite{brameier2023artificial} &       & 2023  &       & Health/ Academia &       & "AI hallucination" is the production of confident responses by an NLP tool that are nonsensical or that seem realistic but are not based on any real-world data. &       & 1 \\
\cmidrule{1-1}\cmidrule{3-3}\cmidrule{5-5}\cmidrule{7-7}\cmidrule{9-9}\cmidrule{11-11}    91    &       & Lee et al. \cite{lee2023benefits} &       & 2023  &       & Health &       & "AI hallucination" refers to the occurrence where GPT-4 produces false responses, which are often presented in a convincing manner, potentially leading the inquirer to believe their accuracy. &       & 260 \\
\cmidrule{1-1}\cmidrule{3-3}\cmidrule{5-5}\cmidrule{7-7}\cmidrule{9-9}\cmidrule{11-11}    92    &       & Long et al. \cite{Long2023.05.30.23290758} &       & 2023  &       & Health/ Academia &       & "AI hallucination" refers to the phenomenon where Language Models (LMs), including ChatGPT, produce outputs characterized by blatant factual errors, significant omissions, and erroneous information generation. &       & 0 \\
\cmidrule{1-1}\cmidrule{3-3}\cmidrule{5-5}\cmidrule{7-7}\cmidrule{9-9}\cmidrule{11-11}    93    &       & Zhang \cite{zhang2023should} &       & 2023  &       & Large Language Model &       & "AI hallucination" refers to the generation of responses by ChatGPT that frequently contain incorrect information, and it can even extend to the generation of fake scientific abstracts and research papers. &       & 1 \\
\cmidrule{1-1}\cmidrule{3-3}\cmidrule{5-5}\cmidrule{7-7}\cmidrule{9-9}\cmidrule{11-11}    94    &       & Puchert et al. \cite{puchert2023llmmaps} &       & 2023  &       & Health &       & "AI hallucination" refers to a common phenomenon where the model includes incorrect or false information in its responses, despite providing eloquent answers. &       & 1 \\
\cmidrule{1-1}\cmidrule{3-3}\cmidrule{5-5}\cmidrule{7-7}\cmidrule{9-9}\cmidrule{11-11}    95    &       & Wang et al. \cite{wang2023chatgpt} &       & 2023  &       & Health &       & "AI hallucination" refers to misinformation, unreasonable or illogical in common-sense knowledge &       & 0 \\
\cmidrule{1-1}\cmidrule{3-3}\cmidrule{5-5}\cmidrule{7-7}\cmidrule{9-9}\cmidrule{11-11}    96    &       & Garg et al. \cite{garg2023exploring} &       & 2023  &       & Health &       & "AI hallucination" refers to the production of content that sounds authoritative but can be inaccurate, incomplete, or biased in nature. &       & 2 \\
\cmidrule{1-1}\cmidrule{3-3}\cmidrule{5-5}\cmidrule{7-7}\cmidrule{9-9}\cmidrule{11-11}    97    &       & Han et al. \cite{han2023large} &       & 2023  &       & Health &       & "AI hallucination" occurs when AI confidently generates an impressive-sounding response that may not be justified by its training data or may even be factually incorrect. &       & 0 \\
\cmidrule{1-1}\cmidrule{3-3}\cmidrule{5-5}\cmidrule{7-7}\cmidrule{9-9}\cmidrule{11-11}    98    &       & Dolan and Freer \cite{dolan5} &       & 2023  &       & Academia &       & "AI hallucination" refers to the phenomenon where AI systems may create or invent text and citations that sound realistic but are not based on actual information, and in certain instances, they may even fabricate nonexistent works. &       & 0 \\
\cmidrule{1-1}\cmidrule{3-3}\cmidrule{5-5}\cmidrule{7-7}\cmidrule{9-9}\cmidrule{11-11}    99    &       & Scott-Branch et al. \cite{scott2023intersection} &       & 2023  &       & Academia &       & "AI hallucination" refers to mistakes in the generated text that are semantically or syntactically plausible but are in fact incorrect or nonsensical. &       & 0 \\
\cmidrule{1-1}\cmidrule{3-3}\cmidrule{5-5}\cmidrule{7-7}\cmidrule{9-9}\cmidrule{11-11}    100   &       & Larsen-Ledet \cite{larsenembracing} &       & 2023  &       & Academia &       & "AI hallucination" refers to the generation of output, often text, containing falsities, which undermine the intended factual nature of the service, intended to be reliable and trustworthy. &       & 0 \\
\cmidrule{1-1}\cmidrule{3-3}\cmidrule{5-5}\cmidrule{7-7}\cmidrule{9-9}\cmidrule{11-11}    101   &       & Kim et al. \cite{kim2023chatgpt} &       & 2023  &       & Health/ Academia &       & "AI hallucination" refers to the phenomenon in which there is no true internal understanding of the language, and LLMs generate text that sounds plausible based on the training data they have been exposed to. &       & 9 \\
\cmidrule{1-1}\cmidrule{3-3}\cmidrule{5-5}\cmidrule{7-7}\cmidrule{9-9}\cmidrule{11-11}    102   &       & Tay et al. \cite{tay2023use} &       & 2023  &       & Health &       & "AI hallucination" refers to the situation where GPT confidently generates responses that lack any justification from training data or correspondence to real-world input. &       & 1 \\
\cmidrule{1-1}\cmidrule{3-3}\cmidrule{5-5}\cmidrule{7-7}\cmidrule{9-9}\cmidrule{11-11}    103   &       & Cai et al. \cite{cai2023performance} &       & 2023  &       & Health &       & "AI hallucination" refers to the issue of accuracy and reliability in LLMs, particularly concerning the high frequency of false information. &       & 9 \\
\cmidrule{1-1}\cmidrule{3-3}\cmidrule{5-5}\cmidrule{7-7}\cmidrule{9-9}\cmidrule{11-11}    104   &       & Jahic et al. \cite{jahic2023harnessing} &       & 2023  &       & Academia &       & "AI hallucination" refers to the phenomenon where an AI system generates unexpected or meaningless outputs that appear coherent and plausible to human observers. &       & 2 \\
\cmidrule{1-1}\cmidrule{3-3}\cmidrule{5-5}\cmidrule{7-7}\cmidrule{9-9}\cmidrule{11-11}    105   &       & Randell and Coghlan \cite{randell2023chatgpt} &       & 2023  &       & Academia &       & "AI hallucination" refers to the plausible yet erroneous text generated by artificial intelligence systems, resembling compression artifacts, which can only be discerned through comparison with original sources like the web or our existing knowledge. &       & 0 \\
\cmidrule{1-1}\cmidrule{3-3}\cmidrule{5-5}\cmidrule{7-7}\cmidrule{9-9}\cmidrule{11-11}    106   &       & Brender \cite{brender2023chatbot} &       & 2023  &       & Academia &       & "AI hallucination" refers to the unfaithful or nonsensical text occasionally generated by large language models. &       & 1 \\
\cmidrule{1-1}\cmidrule{3-3}\cmidrule{5-5}\cmidrule{7-7}\cmidrule{9-9}\cmidrule{11-11}    107   &       & Pedersen \cite{pedersen2023adapting} &       & 2023  &       & Academia &       & "AI hallucination" refers to mistakes in the generated text that are semantically or syntactically plausible but are in fact incorrect or nonsensical. &       & 0 \\
\cmidrule{1-1}\cmidrule{3-3}\cmidrule{5-5}\cmidrule{7-7}\cmidrule{9-9}\cmidrule{11-11}    108   &       & Munoz et al. \cite{munoz2023aain} &       & 2023  &       & Academia &       & “AI hallucination” which results when the system provides a response that is not factual. &       & 1 \\
\cmidrule{1-1}\cmidrule{3-3}\cmidrule{5-5}\cmidrule{7-7}\cmidrule{9-9}\cmidrule{11-11}    109   &       & Hatem et al. \cite{hatem2023call} &       & 2023  &       & Health &       & "AI hallucination" is inaccurate and stigmatizing to both AI systems and individuals who experience hallucinations. Because of this, they suggest the alternative term "AI misinformation" as they feel this is an appropriate term to describe the phenomenon at hand without attributing lifelike characteristics to AI. &       & 0 \\
\cmidrule{1-1}\cmidrule{3-3}\cmidrule{5-5}\cmidrule{7-7}\cmidrule{9-9}\cmidrule{11-11}    110   &       & Athaluri et al. \cite{athaluri2023exploring} &       & 2023  &       & Health/ Academia &       & "AI hallucination" is a phenomenon where AI generates a convincing but completely made-up answer. &       & 16 \\
\cmidrule{1-1}\cmidrule{3-3}\cmidrule{5-5}\cmidrule{7-7}\cmidrule{9-9}\cmidrule{11-11}    111   &       & Gorichanaz \cite{gorichanaz2023accused} &       & 2023  &       & Academia &       & "AI hallucination" refers to generating misinformation, including making false statements, citing sources that do not exist and creating code that doesn’t work. &       & 1 \\
\cmidrule{1-1}\cmidrule{3-3}\cmidrule{5-5}\cmidrule{7-7}\cmidrule{9-9}\cmidrule{11-11}    112   &       & Cox et al. \cite{cox2023prompting} &       & 2023  &       & Health &       & "AI hallucination" occurs when Language Models (LLMs) misunderstand medical vocabulary or provide advice that does not align with established medical guidelines. &       & 0 \\
\cmidrule{1-1}\cmidrule{3-3}\cmidrule{5-5}\cmidrule{7-7}\cmidrule{9-9}\cmidrule{11-11}    113   &       & Yadava \cite{yadava2023chatgpt} &       & 2023  &       & Health/ Academia &       & "AI hallucination" is interestingly labeled as a phenomenon where responses from ChatGPT are sometimes ambiguous, nonsensical, and undesirable. &       & 11 \\
\cmidrule{1-1}\cmidrule{3-3}\cmidrule{5-5}\cmidrule{7-7}\cmidrule{9-9}\cmidrule{11-11}    114   &       & Dai et al. \cite{dai2023reconceptualizing} &       & 2023  &       & Academia &       & "AI hallucination" is a phenomenon in which it sometimes produces confident but irrelevant or inaccurate responses. &       & 11 \\
\cmidrule{1-1}\cmidrule{3-3}\cmidrule{5-5}\cmidrule{7-7}\cmidrule{9-9}\cmidrule{11-11}    115   &       & Lingard \cite{lingard2023writing} &       & 2023  &       & Academia &       & "AI hallucination" refers to the phenomenon where AI generates content that confidently presents as legitimate-sounding material but is not real. &       & 1 \\
\cmidrule{1-1}\cmidrule{3-3}\cmidrule{5-5}\cmidrule{7-7}\cmidrule{9-9}\cmidrule{11-11}    116   &       & Woodland \cite{woodland2023chatgpt} &       & 2023  &       & Health/ Academia &       & "AI hallucination" refers to a situation where the generated text is not derived from a factual source but instead arises from statistical predictions of words that are likely to follow the given input. &       & 0 \\
\cmidrule{1-1}\cmidrule{3-3}\cmidrule{5-5}\cmidrule{7-7}\cmidrule{9-9}\cmidrule{11-11}    117   &       & Kashangura \cite{kashangura2023human} &       & 2023  &       & Health &       & "AI hallucination" is the generation or production of a factually invalid statement, characterized by confident output from AI that is not correct or not real. &       & 0 \\
\cmidrule{1-1}\cmidrule{3-3}\cmidrule{5-5}\cmidrule{7-7}\cmidrule{9-9}\cmidrule{11-11}    118   &       & Walker et al. \cite{walker2023reliability} &       & 2023  &       & Health &       & “AI hallucinations” known as wrong or out of context answers generated by ChatGPT AI. &       & 6 \\
\cmidrule{1-1}\cmidrule{3-3}\cmidrule{5-5}\cmidrule{7-7}\cmidrule{9-9}\cmidrule{11-11}    119   &       & Tenge Hansen and R{\o}sand Val{\o} \cite{tenge2023cybersecurity} &       & 2023  &       & Academia &       & “AI hallucination” is defined as a phenomenon where AI generates a convincing but completely made-up answer, often incorporating fake references for added persuasiveness.\newline{}“Intrinsic hallucination” refers to the LLM generation that contradicts the source/input.\newline{}“Extrinsic hallucination” refers to the LLM generations that cannot be verified from the source/input content, i.e., output that can neither be supported nor contradicted by the source. &       & 0 \\
\cmidrule{1-1}\cmidrule{3-3}\cmidrule{5-5}\cmidrule{7-7}\cmidrule{9-9}\cmidrule{11-11}    120   &       & Tupper et al. \cite{tupper2023field} &       & 2023  &       & Academia &       & "AI hallucination" refers to a phenomenon wherein incorrect or nonsensical responses are generated in response to prompts. &       & 0 \\
\cmidrule{1-1}\cmidrule{3-3}\cmidrule{5-5}\cmidrule{7-7}\cmidrule{9-9}\cmidrule{11-11}    121   &       & Borkowski et al. \cite{borkowski2023applications} &       & 2023  &       & Health &       & "AI Hallucination" refers to the phenomenon often observed in language models like ChatGPT, characterized by the generation of inaccurate information, which result from the model's inability to differentiate between real and fake information sources. &       & 0 \\
\cmidrule{1-1}\cmidrule{3-3}\cmidrule{5-5}\cmidrule{7-7}\cmidrule{9-9}\cmidrule{11-11}    122   &       & Subramanya and Furlong \cite{subramanya2023revolutionising} &       & 2023  &       & Legal Setting &       & "AI hallucination" occurs when AI generates content that appears accurate but includes false references. &       & 0 \\
\cmidrule{1-1}\cmidrule{3-3}\cmidrule{5-5}\cmidrule{7-7}\cmidrule{9-9}\cmidrule{11-11}    123   &       & Tenzer \cite{tenzer2023defamation} &       & 2023  &       & Legal Setting &       & "AI hallucination" occurs when a chatbot produces a confident but inaccurate response to a question, contributing to the erratic and unreliable behavior of A.I. large language models (L.L.M.s), which may include providing false information and acting strangely toward users. &       & 0 \\
\cmidrule{1-1}\cmidrule{3-3}\cmidrule{5-5}\cmidrule{7-7}\cmidrule{9-9}\cmidrule{11-11}    124   &       & Alarie and McCreight \cite{alarie2023ethics} &       & 2023  &       & Legal Setting &       & "AI hallucination" occurs when AI generates untrue information that is not backed up by real-world data. &       & 0 \\
\cmidrule{1-1}\cmidrule{3-3}\cmidrule{5-5}\cmidrule{7-7}\cmidrule{9-9}\cmidrule{11-11}    125   &       & Han et al. \cite{han2023does} &       & 2023  &       & Academia &       & "AI hallucination" refers to the situation where the AI, while not presenting fictitious information like a hallucination, draws from sources other than the requested one and fails to indicate this, potentially leading to incorrect conclusions when users rely on the AI-generated text for understanding a specific paper. &       & 0 \\
\cmidrule{1-1}\cmidrule{3-3}\cmidrule{5-5}\cmidrule{7-7}\cmidrule{9-9}\cmidrule{11-11}    126   &       & Picht \cite{picht2023chatgpt} &       & 2023  &       & Legal Setting &       & "AI Hallucination" refers to the phenomenon in which ChatGPT generates responses that sound confident and compelling, despite its inability to genuinely answer the posed question. &       & 0 \\
\cmidrule{1-1}\cmidrule{3-3}\cmidrule{5-5}\cmidrule{7-7}\cmidrule{9-9}\cmidrule{11-11}    127   &       & Treleaven et al. \cite{treleaven2023} &       & 2023  &       & Legal Setting &       & "AI hallucination" is defined as a confident response that is biased, too specialized, or even entirely incorrect, with the model fabricating a seemingly plausible but factually inaccurate answer. &       & 0 \\
\cmidrule{1-1}\cmidrule{3-3}\cmidrule{5-5}\cmidrule{7-7}\cmidrule{9-9}\cmidrule{11-11}    128   &       & Ariyaratne \cite{ariyaratne2023} &       & 2023  &       & Legal Setting &       & "AI Hallucination" refers to the term used to describe the phenomena of generative algorithms making up facts. &       & 0 \\
\cmidrule{1-1}\cmidrule{3-3}\cmidrule{5-5}\cmidrule{7-7}\cmidrule{9-9}\cmidrule{11-11}    129   &       & Wu et al. \cite{wu2023brief} &       & 2023  &       & Large Language Model &       & "AI hallucination" occurs when the replies generated by ChatGPT frequently contain factual errors. &       & 40 \\
\cmidrule{1-1}\cmidrule{3-3}\cmidrule{5-5}\cmidrule{7-7}\cmidrule{9-9}\cmidrule{11-11}    130   &       & Wang et al. \cite{wang2023survey} &       & 2023  &       & Large Language Model &       & "AI hallucination" occurs when ChatGPT produces responses that, despite sounding plausible, are ultimately incorrect or nonsensical. &       & 7 \\
\cmidrule{1-1}\cmidrule{3-3}\cmidrule{5-5}\cmidrule{7-7}\cmidrule{9-9}\cmidrule{11-11}    131   &       & Amaro et al. \cite{amaro2023believe} &       & 2023  &       & Large Language Model &       & "AI hallucination" occurs when ChatGPT generates outputs that invent facts and concepts, as it lacks objective training to assess the factual correctness of its responses. &       & 0 \\
\cmidrule{1-1}\cmidrule{3-3}\cmidrule{5-5}\cmidrule{7-7}\cmidrule{9-9}\cmidrule{11-11}    132   &       & Skrodelis et al. \cite{skrodelis2023latest} &       & 2023  &       & Natural Language Generation &       & "AI hallucination" occurs when NLGs frequently produce text that is either nonsensical or not true to the original input. &       & 0 \\
\cmidrule{1-1}\cmidrule{3-3}\cmidrule{5-5}\cmidrule{7-7}\cmidrule{9-9}\cmidrule{11-11}    133   &       & Gupta et al. \cite{gupta2023chatgpt} &       & 2023  &       & Cybersecurity &       & "AI hallucination" refers to a phenomenon in which the AI model generates inaccurate or outright false information. &       & 7 \\
\cmidrule{1-1}\cmidrule{3-3}\cmidrule{5-5}\cmidrule{7-7}\cmidrule{9-9}\cmidrule{11-11}    134   &       & Bahrini et al. \cite{bahrini2023chatgpt} &       & 2023  &       & Chatbot &       & "AI hallucination" refers to ChatGPT's limitations and challenges, which encompass bias and the occasional generation of nonsensical output &       & 28 \\
\cmidrule{1-1}\cmidrule{3-3}\cmidrule{5-5}\cmidrule{7-7}\cmidrule{9-9}\cmidrule{11-11}    135   &       & Hamid \cite{hamid2023chatgpt} &       & 2023  &       & Chatbot &       & "AI hallucination" refers to the 'hallucinations,' or incoherent responses, observed in Language Models (LLMs), which are essentially errors in the model's output as it is designed to prioritize coherence rather than truth. &       & 1 \\
\cmidrule{1-1}\cmidrule{3-3}\cmidrule{5-5}\cmidrule{7-7}\cmidrule{9-9}\cmidrule{11-11}    136   &       & De Silva et al. \cite{de2023chatgpt} &       & 2023  &       & Chatbot &       & "AI hallucination" occurs when ChatGPT exhibits anomalous behavior, producing factual inaccuracies, logical fallacies, bias, and plagiarism in its responses, a phenomenon popularly referred to as 'AI hallucinations' or 'stochastic parroting.' &       & 3 \\
\cmidrule{1-1}\cmidrule{3-3}\cmidrule{5-5}\cmidrule{7-7}\cmidrule{9-9}\cmidrule{11-11}    137   &       & Atallah \cite{atallah2023large} &       & 2023  &       & Health &       & "AI hallucination" occurs when modern LLMs produce fluent and grammatically correct textual outputs that are categorically false, and in some instances, fictitious. &       & 1 \\
\cmidrule{1-1}\cmidrule{3-3}\cmidrule{5-5}\cmidrule{7-7}\cmidrule{9-9}\cmidrule{11-11}    138   &       & Byrne \cite{byrne2023disruptive} &       & 2023  &       & Large Language Model &       & "AI hallucination" refers to outputs that reflect misinterpretations and falsehoods. &       & 0 \\
\cmidrule{1-1}\cmidrule{3-3}\cmidrule{5-5}\cmidrule{7-7}\cmidrule{9-9}\cmidrule{11-11}    139   &       & Thirunavukarasu et al. \cite{thirunavukarasu2023large} &       & 2023  &       & Health &       & "AI hallucination" refers to the occurrence where inaccurate information is invented, not represented in the training dataset, and is presented lucidly, with an alternative term like 'fact fabrication' being preferred. &       & 30 \\
\cmidrule{1-1}\cmidrule{3-3}\cmidrule{5-5}\cmidrule{7-7}\cmidrule{9-9}\cmidrule{11-11}    140   &       & Ting et al. \cite{ting2023chatgpt} &       & 2023  &       & Health &       & AI Hallucination refers to the phenomenon where, as a generative AI, ChatGPT generates outputs based on statistical prediction of the text without human-like reasoning, potentially resulting in plausible-sounding but inaccurate responses, also known as "fabrication." &       & 5 \\
\cmidrule{1-1}\cmidrule{3-3}\cmidrule{5-5}\cmidrule{7-7}\cmidrule{9-9}\cmidrule{11-11}    141   &       & Beam et al. \cite{beam2023artificial} &       & 2023  &       & Health &       & "AI hallucination" occurs when a model generates information that isn't present in the input data, leading to outputs that seem plausible but are factually incorrect or nonsensical. &       & 0 \\
\cmidrule{1-1}\cmidrule{3-3}\cmidrule{5-5}\cmidrule{7-7}\cmidrule{9-9}\cmidrule{11-11}    142   &       & Madden et al. \cite{madden2023assessing} &       & 2023  &       & Health &       & "AI hallucination" occurs when ChatGPT generates false information, also called "delusions." &       & 2 \\
\cmidrule{1-1}\cmidrule{3-3}\cmidrule{5-5}\cmidrule{7-7}\cmidrule{9-9}\cmidrule{11-11}    143   &       & Komorowski \cite{komorowski2023could} &       & 2023  &       & Health &       & "AI hallucination" refers to the phenomenon where ChatGPT has the ability to confidently produce answers that appear believable but may be incorrect or nonsensical. &       & 4 \\
\cmidrule{1-1}\cmidrule{3-3}\cmidrule{5-5}\cmidrule{7-7}\cmidrule{9-9}\cmidrule{11-11}    144   &       & Sallam \cite{sallam2023chatgpt} &       & 2023  &       & Health/ Academia &       & "AI hallucination" refers to concerns arising from possible bias in ChatGPT's training datasets, limiting its capabilities and potentially causing factual inaccuracies that, surprisingly, seem scientifically plausible. &       & 334 \\
\cmidrule{1-1}\cmidrule{3-3}\cmidrule{5-5}\cmidrule{7-7}\cmidrule{9-9}\cmidrule{11-11}    145   &       & Rothschild \cite{rothschild2023artificial} &       & 2023  &       & Health/ Academia &       & "AI hallucination" refers to false responses provided by ChatGPT3. &       & 0 \\
\cmidrule{1-1}\cmidrule{3-3}\cmidrule{5-5}\cmidrule{7-7}\cmidrule{9-9}\cmidrule{11-11}    146   &       & Im \cite{im2023} &       & 2023  &       & Health &       & "AI hallucination" occurs when ChatGPT generates responses that are unfaithful to the provided training data. &       & 0 \\
\cmidrule{1-1}\cmidrule{3-3}\cmidrule{5-5}\cmidrule{7-7}\cmidrule{9-9}\cmidrule{11-11}    147   &       & Hulman et al. \cite{hulman2023chatgpt} &       & 2023  &       & Health &       & "AI hallucination" refers to the occurrence of making up completely false information supported by fictitious citations. &       & 5 \\
\cmidrule{1-1}\cmidrule{3-3}\cmidrule{5-5}\cmidrule{7-7}\cmidrule{9-9}\cmidrule{11-11}    148   &       & Cheung et al. \cite{cheung2023chatgpt} &       & 2023  &       & Health &       & "AI hallucination" occurs as a phenomenon where the AI generates responses that are nonsensical or unfaithful to the provided source input. &       & 0 \\
\cmidrule{1-1}\cmidrule{3-3}\cmidrule{5-5}\cmidrule{7-7}\cmidrule{9-9}\cmidrule{11-11}    149   &       & Moskatel and Zhang \cite{moskatel2023utility} &       & 2023  &       & Academia &       & "AI hallucination" is an inaccurate response by an AI that is not justified by its training data. &       & 0 \\
\cmidrule{1-1}\cmidrule{3-3}\cmidrule{5-5}\cmidrule{7-7}\cmidrule{9-9}\cmidrule{11-11}    150   &       & Javid et al. \cite{javid2023emergence} &       & 2023  &       & Health/ Academia &       & "AI hallucination" refers to the phenomenon where the model produces text that is either factually incorrect or nonsensical. &       & 0 \\
\cmidrule{1-1}\cmidrule{3-3}\cmidrule{5-5}\cmidrule{7-7}\cmidrule{9-9}\cmidrule{11-11}    151   &       & Buholayka et al. \cite{buholayka2023readiness} &       & 2023  &       & Health/ Academia &       & "AI hallucination" occurs when AI-generated content is nonsensical or unfaithful to the provided source content. &       & 4 \\
\cmidrule{1-1}\cmidrule{3-3}\cmidrule{5-5}\cmidrule{7-7}\cmidrule{9-9}\cmidrule{11-11}    152   &       & Alkaissi and McFarlane \cite{alkaissi2023artificial} &       & 2023  &       & Health/ Academia &       & "AI hallucination" occurs when ChatGPT provides confident responses that appear faithful but are nonsensical when evaluated against common knowledge. &       & 230 \\
\cmidrule{1-1}\cmidrule{3-3}\cmidrule{5-5}\cmidrule{7-7}\cmidrule{9-9}\cmidrule{11-11}    153   &       & Eysenbach \cite{eysenbach2023role} &       & 2023  &       & Health/ Academia &       & "AI hallucination" is a confident response by an artificial intelligence system that does not seem to be justified by its training data. &       & 128 \\
\cmidrule{1-1}\cmidrule{3-3}\cmidrule{5-5}\cmidrule{7-7}\cmidrule{9-9}\cmidrule{11-11}    154   &       & {\O}stergaard and Nielbo \cite{ostergaard2023false} &       & 2023  &       & Health &       & "Non sequitur", Latin for “it does not follow,” is a term commonly used in philosophy and rhetoric to describe inferences not following from the premises. \newline{}They acknowledged that “non sequitur” does not cover all false responses generated by AI models. Indeed, AI models can also make “hasty generalizations,” i.e., the fallacy of making (too) strong claims based on (too) limited data. \newline{}They realized that many other terms from philosophy and rhetoric describing logical fallacies of different types (e.g., “false analogy,” “appeal to authority” and “false dilemma”) could also be used to label false responses from AI models. &       & 0 \\
\cmidrule{1-1}\cmidrule{3-3}\cmidrule{5-5}\cmidrule{7-7}\cmidrule{9-9}\cmidrule{11-11}    155   &       & Thorne \cite{thorne2023experimenting} &       & 2023  &       & Academia &       & "AI hallucination" refers to occurrences where ChatGPT produces statements that are difficult to verify, appear plausible, but do not withstand scrutiny, often being arbitrary in nature. &       & 0 \\
\cmidrule{1-1}\cmidrule{3-3}\cmidrule{5-5}\cmidrule{7-7}\cmidrule{9-9}\cmidrule{11-11}    156   &       & Huang et al. \cite{huang2023benchmarking} &       & 2023  &       & Health/ Academia &       & "AI hallucination" refers to responses generated by LLMs in a convincing appearance but are actually incorrect statements. &       & 5 \\
\cmidrule{1-1}\cmidrule{3-3}\cmidrule{5-5}\cmidrule{7-7}\cmidrule{9-9}\cmidrule{11-11}    157   &       & Wecel et al. \cite{wecel2023} &       & 2023  &       & Large Language Model &       & "AI hallucination" refers to the occurrence where AI produces texts that are not consistent with reality and contain confused facts. &       & 0 \\
\cmidrule{1-1}\cmidrule{3-3}\cmidrule{5-5}\cmidrule{7-7}\cmidrule{9-9}\cmidrule{11-11}    158   &       & Ge and Lai \cite{ge2023} &       & 2023  &       & Health/ Academia &       & "AI hallucination" occurs where the LLM model generates confident, specific, and fluent answers that are factually completely wrong. Also called "stochastic parroting." &       & 21 \\
\cmidrule{1-1}\cmidrule{3-3}\cmidrule{5-5}\cmidrule{7-7}\cmidrule{9-9}\cmidrule{11-11}    159   &       & Bhatti \cite{bhatti2023} &       & 2023  &       & Health/ Academia &       & "AI hallucination" occurs when the information generated by ChatGPT may not always be correct. &       & 0 \\
\cmidrule{1-1}\cmidrule{3-3}\cmidrule{5-5}\cmidrule{7-7}\cmidrule{9-9}\cmidrule{11-11}    160   &       & Bryant \cite{bryant2023} &       & 2023  &       & Academia &       & "AI hallucination" refers to answers that are fabricated when data are insufficient for an accurate response. &       & 2 \\
\cmidrule{1-1}\cmidrule{3-3}\cmidrule{5-5}\cmidrule{7-7}\cmidrule{9-9}\cmidrule{11-11}    161   &       & Mahyoob et al. \cite{mahyoob2023} &       & 2023  &       & Academia &       & "AI hallucination" occurs when it generates ideas with human-like fluency and persuasiveness but without truth or factual accuracy. &       & 0 \\
\cmidrule{1-1}\cmidrule{3-3}\cmidrule{5-5}\cmidrule{7-7}\cmidrule{9-9}\cmidrule{11-11}    162   &       & Lee \cite{lee2023mathematical} &       & 2023  &       & Mathematical Setting &       & "AI hallucination" occurs when GPT models generate outputs that are contextually implausible or inconsistent with the real world. &       & 14 \\
\cmidrule{1-1}\cmidrule{3-3}\cmidrule{5-5}\cmidrule{7-7}\cmidrule{9-9}\cmidrule{11-11}    163   &       & Pi{\~n}eiro-Mart{\'\i}n et al. \cite{pineiro2023ethical} &       & 2023  &       & Ethical Setting &       & "AI hallucination" occurs when the LLM generates text that goes beyond the scope of the provided input or fabricates information that is factually incorrect. &       & 0 \\
\cmidrule{1-1}\cmidrule{3-3}\cmidrule{5-5}\cmidrule{7-7}\cmidrule{9-9}\cmidrule{11-11}    164   &       & Alhaidry et al. \cite{alhaidry2023chatgpt} &       & 2023  &       & Health &       & "AI hallucination" occurs when ChatGPT can produce answers that appear reliable but are incorrect. &       & 10 \\
\cmidrule{1-1}\cmidrule{3-3}\cmidrule{5-5}\cmidrule{7-7}\cmidrule{9-9}\cmidrule{11-11}    165   &       & Khurana and Vaddi \cite{khurana2023chatgpt} &       & 2023  &       & Health/ Academia &       & "AI hallucination" occurs when AI generates sentences with the intent to convince the reader, which can be misleading, particularly to inexpert readers. &       & 1 \\
\cmidrule{1-1}\cmidrule{3-3}\cmidrule{5-5}\cmidrule{7-7}\cmidrule{9-9}\cmidrule{11-11}    166   &       & Li et al. \cite{li2023chatdoctor} &       & 2023  &       & Health &       & "AI hallucination" occurs when LLMs occasionally generate fallacious and harmful assertions beyond their knowledge expertise. &       & 11 \\
\cmidrule{1-1}\cmidrule{3-3}\cmidrule{5-5}\cmidrule{7-7}\cmidrule{9-9}\cmidrule{11-11}    167   &       & Walczak and Cellary \cite{walczak2023challenges} &       & 2023  &       & Academia &       & "AI hallucination" occurs as a primary limitation associated with LLMs, manifesting as the tendency to generate errors, including mathematical, computational, and conceptual inaccuracies, without any prior indication, often characterized by their deceptive plausibility, alignment with truthful information, and conveyed in a persuasive and self-assured manner, making their detection challenging without careful scrutiny and rigorous fact verification. &       & 0 \\
\cmidrule{1-1}\cmidrule{3-3}\cmidrule{5-5}\cmidrule{7-7}\cmidrule{9-9}\cmidrule{11-11}    168   &       & Marinaccio et al. \cite{marinaccio2023} &       & 2023  &       & Legal Setting &       & "AI hallucination" occurs when the output is delivered in an authoritative and convincing manner, potentially leading uninformed users to blindly accept it as truth, despite its inaccuracies or falsehoods. &       & 0 \\
\cmidrule{1-1}\cmidrule{3-3}\cmidrule{5-5}\cmidrule{7-7}\cmidrule{9-9}\cmidrule{11-11}    169   &       & Gebrael et al. \cite{gebrael2023enhancing} &       & 2023  &       & Health &       & "AI hallucination" occurs when AI models, particularly in cases like ChatGPT, generate outputs that appear plausible but are factually incorrect or unrelated to the input context. &       & 3 \\
\cmidrule{1-1}\cmidrule{3-3}\cmidrule{5-5}\cmidrule{7-7}\cmidrule{9-9}\cmidrule{11-11}    170   &       & Tiwari et al. \cite{tiwari2023implications} &       & 2023  &       & Health &       & "AI hallucination" occurs as false knowledge that appears convincing from a scientific perspective. &       & 5 \\
\cmidrule{1-1}\cmidrule{3-3}\cmidrule{5-5}\cmidrule{7-7}\cmidrule{9-9}\cmidrule{11-11}    171   &       & Bhattacharyya \cite{bhattacharyya2023high} &       & 2023  &       & Health/ Academia &       & "AI hallucination" occurs as a phenomenon where nonsensical or inaccurate content is generated. &       & 10 \\
\cmidrule{1-1}\cmidrule{3-3}\cmidrule{5-5}\cmidrule{7-7}\cmidrule{9-9}\cmidrule{11-11}    172   &       & Sriwastwa et al. \cite{sriwastwa2023generative} &       & 2023  &       & Academia &       & "AI hallucination" refers to the phenomenon in ChatGPT output where the text is cogent but not necessarily true, often presenting as a complete fabrication. &       & 0 \\
\cmidrule{1-1}\cmidrule{3-3}\cmidrule{5-5}\cmidrule{7-7}\cmidrule{9-9}\cmidrule{11-11}    173   &       & Dossantos et al. \cite{dossantos2023eyes} &       & 2023  &       & Health &       & "AI hallucination" occurs when ChatGPT produces inaccurate results, particularly in specialized topics, due to a lack of depth and inaccurate details retrieved from the LLM's database, with no assurance that ChatGPT's suggestions adhere to evidence-based guidelines or best practices. &       & 1 \\
\cmidrule{1-1}\cmidrule{3-3}\cmidrule{5-5}\cmidrule{7-7}\cmidrule{9-9}\cmidrule{11-11}    174   &       & Loos et al. \cite{loos2023using} &       & 2023  &       & Academia &       & "AI hallucination" occurs as a phenomenon characterized by the generation of incorrect or outdated information, accompanied by the failure to provide reliable sources to evaluate the generated content. &       & 0 \\
\cmidrule{1-1}\cmidrule{3-3}\cmidrule{5-5}\cmidrule{7-7}\cmidrule{9-9}\cmidrule{11-11}    175   &       & Koga \cite{koga2023integration} &       & 2023  &       & Academia &       & "AI hallucination" occurs when LLMs generate seemingly credible but fabricated information, particularly concerning when they create fictitious citations. &       & 0 \\
\cmidrule{1-1}\cmidrule{3-3}\cmidrule{5-5}\cmidrule{7-7}\cmidrule{9-9}\cmidrule{11-11}    176   &       & Birenbaum \cite{birenbaum2023chatbots} &       & 2023  &       & Academia &       & "AI hallucination" occurs when the model fabricates information and provides untraceable references to support its claims. &       & 3 \\
\cmidrule{1-1}\cmidrule{3-3}\cmidrule{5-5}\cmidrule{7-7}\cmidrule{9-9}\cmidrule{11-11}    177   &       & Cascella et al. \cite{cascella2023evaluating} &       & 2023  &       & Health &       & "AI hallucination" refers to the ability of ChatGPT to produce answers that sound believable but may be incorrect or nonsensical. &       & 182 \\
\cmidrule{1-1}\cmidrule{3-3}\cmidrule{5-5}\cmidrule{7-7}\cmidrule{9-9}\cmidrule{11-11}    178   &       & Aronson \cite{aronson2023use} &       & 2023  &       & Health/ Academia &       & "AI hallucination" refers to all the pieces of misinformation generated. &       & 0 \\
\cmidrule{1-1}\cmidrule{3-3}\cmidrule{5-5}\cmidrule{7-7}\cmidrule{9-9}\cmidrule{11-11}    179   &       & Kumar et al. \cite{kumar2023artificial} &       & 2023  &       & Academia &       & "AI hallucination" refers to instances where an AI chatbot generates fictional, erroneous, or unsubstantiated information in response to queries. &       & 0 \\
\cmidrule{1-1}\cmidrule{3-3}\cmidrule{5-5}\cmidrule{7-7}\cmidrule{9-9}\cmidrule{11-11}    180   &       & Gravel et al. \cite{gravel2023learning} &       & 2023  &       & Academia &       & ChatGPT fabricated a convincing response that contained several factual errors. &       & 19 \\
\cmidrule{1-1}\cmidrule{3-3}\cmidrule{5-5}\cmidrule{7-7}\cmidrule{9-9}\cmidrule{11-11}    181   &       & Ferres et al. \cite{ferres2023} &       & 2023  &       & Health &       & "AI hallucination" occurs when a model generates content that has no basis in reality, creating entirely made-up stories or facts. &       & 8 \\
\cmidrule{1-1}\cmidrule{3-3}\cmidrule{5-5}\cmidrule{7-7}\cmidrule{9-9}\cmidrule{11-11}    182   &       & Varghese and Chapiro \cite{varghese2023} &       & 2023  &       & Health &       & "AI hallucination" occurs when faced with topics that the model has not received adequate training or supervision for, resulting in fabricated output delivered with a strong sense of certainty. &       & 1 \\
\cmidrule{1-1}\cmidrule{3-3}\cmidrule{5-5}\cmidrule{7-7}\cmidrule{9-9}\cmidrule{11-11}    183   &       & Dunn and Cianflone \cite{dunn2023} &       & 2023  &       & Health &       & "AI hallucination" occurs as a limitation, characterized by the lack of transparency in how the output is generated. &       & 1 \\
\cmidrule{1-1}\cmidrule{3-3}\cmidrule{5-5}\cmidrule{7-7}\cmidrule{9-9}\cmidrule{11-11}    184   &       & Theodosiou and Read \cite{theodosiou2023} &       & 2023  &       & Health &       & "AI hallucinations" is the phenomenon in which a generative AI tool confidently asserts a factual inaccuracy. &       & 5 \\
\cmidrule{1-1}\cmidrule{3-3}\cmidrule{5-5}\cmidrule{7-7}\cmidrule{9-9}\cmidrule{11-11}    185   &       & Fesenmaier and Wober \cite{FESENMAIER2023103578} &       & 2023  &       & Academia &       & "AI hallucination" occurs when ChatGPT appears to create fake or inaccurate findings that seem plausible. &       & 0 \\
\cmidrule{1-1}\cmidrule{3-3}\cmidrule{5-5}\cmidrule{7-7}\cmidrule{9-9}\cmidrule{11-11}    186   &       & Sohail et al. \cite{sohail2023decoding} &       & 2023  &       & ChatGPT &       & "AI hallucination" occurs when ChatGPT generates new data or information that does not exist. &       & 5 \\
\cmidrule{1-1}\cmidrule{3-3}\cmidrule{5-5}\cmidrule{7-7}\cmidrule{9-9}\cmidrule{11-11}    187   &       & Karakas et al. \cite{karakas4475000leveraging} &       & 2023  &       & Health &       & "AI hallucination" occurs when the AI sometimes generates responses that sound plausible but are, in fact, incorrect. Also called "confabulation." &       & 0 \\
\cmidrule{1-1}\cmidrule{3-3}\cmidrule{5-5}\cmidrule{7-7}\cmidrule{9-9}\cmidrule{11-11}    188   &       & Lyons et al. \cite{lyons2023artificial} &       & 2023  &       & Health &       & "AI hallucination" occurs when LLMs create responses to prompts by sampling from the language distribution within their dataset, potentially leading to the generation of incorrect information and the propagation of biases. &       & 0 \\
\cmidrule{1-1}\cmidrule{3-3}\cmidrule{5-5}\cmidrule{7-7}\cmidrule{9-9}\cmidrule{11-11}    189   &       & Opdahl et al. \cite{opdahl2023trustworthy} &       & 2023  &       & Journalism &       & "AI hallucination" occurs when the model generates plausible-sounding nonsense, including texts that contain elements not found in the input data. &       & 1 \\
\cmidrule{1-1}\cmidrule{3-3}\cmidrule{5-5}\cmidrule{7-7}\cmidrule{9-9}\cmidrule{11-11}    190   &       & Blanchard et al. \cite{blanchard2023chatgpt} &       & 2023  &       & Health/ Academia &       & "AI hallucination" occurs when probabilities from the transformer model architecture can generate made-up answers, taking numerous forms, from false references in scientific reports to misinformation in journalistic articles or incorrect formulas when debugging programming code. &       & 2 \\
\cmidrule{1-1}\cmidrule{3-3}\cmidrule{5-5}\cmidrule{7-7}\cmidrule{9-9}\cmidrule{11-11}    191   &       & Lim et al. \cite{lim2023benchmarking} &       & 2023  &       & Health &       & "AI hallucination" occurs when LLMs lack domain-specific capabilities, rendering them susceptible to generating convincing yet potentially inaccurate responses. &       & 1 \\
\cmidrule{1-1}\cmidrule{3-3}\cmidrule{5-5}\cmidrule{7-7}\cmidrule{9-9}\cmidrule{11-11}    192   &       & Kim et al. \cite{kim2023ai} &       & 2023  &       & Health &       & "AI hallucination" occurs as the possibility of providing incorrect information or exhibiting errors in the inference process. &       & 0 \\
\cmidrule{1-1}\cmidrule{3-3}\cmidrule{5-5}\cmidrule{7-7}\cmidrule{9-9}\cmidrule{11-11}    193   &       & Alqahtani et al. \cite{alqahtani2023emergent} &       & 2023  &       & Academia &       & "AI hallucination" refers to generating non-existent or incorrect content and other related concerns associated with limited contexts, reliability, and the lack of learning from experience. &       & 15 \\
\cmidrule{1-1}\cmidrule{3-3}\cmidrule{5-5}\cmidrule{7-7}\cmidrule{9-9}\cmidrule{11-11}    194   &       & Jairoun et al. \cite{jairoun2023chatgpt} &       & 2023  &       & Health &       & "AI hallucination" occurs as statements that are factually inaccurate yet appear plausible to the layman. &       & 3 \\
\cmidrule{1-1}\cmidrule{3-3}\cmidrule{5-5}\cmidrule{7-7}\cmidrule{9-9}\cmidrule{11-11}    195   &       & {\v{S}}lapeta \cite{vslapeta2023chatgpt} &       & 2023  &       & Health &       & "AI hallucination" occurs when highly confident answers are returned by AI that cannot be explained by the training data alone. &       & 12 \\
\cmidrule{1-1}\cmidrule{3-3}\cmidrule{5-5}\cmidrule{7-7}\cmidrule{9-9}\cmidrule{11-11}    196   &       & Dillion et al. \cite{dillion2023can} &       & 2023  &       & Science &       & "AI hallucination" occurs as outputs that appear sensical but are inaccurate. &       & 28 \\
\cmidrule{1-1}\cmidrule{3-3}\cmidrule{5-5}\cmidrule{7-7}\cmidrule{9-9}\cmidrule{11-11}    197   &       & Thomson et al. \cite{thomson2023evaluating} &       & 2023  &       & Text Summarization &       & "AI hallucination" occurs when the output text includes an attribute that was not present in the input data. &       & 3 \\
\cmidrule{1-1}\cmidrule{3-3}\cmidrule{5-5}\cmidrule{7-7}\cmidrule{9-9}\cmidrule{11-11}    198   &       & Balas and Ing \cite{balas2023conversational} &       & 2023  &       & Health &       & "AI hallucination" refers to producing confident responses that sound plausible yet are factually incorrect. &       & 19 \\
\cmidrule{1-1}\cmidrule{3-3}\cmidrule{5-5}\cmidrule{7-7}\cmidrule{9-9}\cmidrule{11-11}    199   &       & Salah et al. \cite{salah2023may} &       & 2023  &       & Science &       & "AI hallucination" refers to information that sounds plausible but is entirely fabricated or not supported by the input data. &       & 2 \\
\cmidrule{1-1}\cmidrule{3-3}\cmidrule{5-5}\cmidrule{7-7}\cmidrule{9-9}\cmidrule{11-11}    200   &       & Ilicki et al. \cite{ilicki2023framework} &       & 2023  &       & Health &       & "AI hallucination" refers to text responses that are either nonsensical or unfaithful to the content they should use. &       & 3 \\
\cmidrule{1-1}\cmidrule{3-3}\cmidrule{5-5}\cmidrule{7-7}\cmidrule{9-9}\cmidrule{11-11}    201   &       & Jansen et al. \cite{jansen2023employing} &       & 2023  &       & Survey Setting &       & "AI hallucination" refers to generating nonsensical or inappropriate responses to survey questions. &       & 2 \\
\cmidrule{1-1}\cmidrule{3-3}\cmidrule{5-5}\cmidrule{7-7}\cmidrule{9-9}\cmidrule{11-11}    202   &       & Casal and Kessler \cite{casal2023can} &       & 2023  &       & Academia &       & "AI hallucination" refers to the tendency to invent content. &       & 1 \\
\cmidrule{1-1}\cmidrule{3-3}\cmidrule{5-5}\cmidrule{7-7}\cmidrule{9-9}\cmidrule{11-11}    203   &       & Qi et al. \cite{qi2023promise} &       & 2023  &       & Health &       & "AI hallucination" occurs when ChatGPT generates responses that appear plausible but require correction, including the invention of terms it is familiar with. &       & 11 \\
\cmidrule{1-1}\cmidrule{3-3}\cmidrule{5-5}\cmidrule{7-7}\cmidrule{9-9}\cmidrule{11-11}    204   &       & Lin \cite{lin2023and} &       & 2023  &       & Academia &       & "AI hallucination" occurs when LLMs fabricate facts, creating confident-sounding statements and legitimate-looking citations that are false. &       & 22 \\
\cmidrule{1-1}\cmidrule{3-3}\cmidrule{5-5}\cmidrule{7-7}\cmidrule{9-9}\cmidrule{11-11}    205   &       & Janssen et al. \cite{janssen2023use} &       & 2023  &       & Health &       & "AI hallucination" occurs where the model generates text that is factually incorrect or nonsensical, despite appearing confident in its ability. &       & 16 \\
\cmidrule{1-1}\cmidrule{3-3}\cmidrule{5-5}\cmidrule{7-7}\cmidrule{9-9}\cmidrule{11-11}    206   &       & Shen et al. \cite{shen2023chatgpt} &       & 2023  &       & Health &       & "AI hallucination" occurs when LLMs produce seemingly credible but incorrect responses, including the invention of terms they should be familiar with. &       & 235 \\
\cmidrule{1-1}\cmidrule{3-3}\cmidrule{5-5}\cmidrule{7-7}\cmidrule{9-9}\cmidrule{11-11}    207   &       & Thirunavukarasu \cite{thirunavukarasu2023trialling} &       & 2023  &       & Health &       & "AI hallucination" occurs when chatGPT describes inaccurate information as lucidly as it does with correct facts. &       & 23 \\
\cmidrule{1-1}\cmidrule{3-3}\cmidrule{5-5}\cmidrule{7-7}\cmidrule{9-9}\cmidrule{11-11}    208   &       & Qadir \cite{qadir2023engineering} &       & 2023  &       & Academia &       & "AI hallucination" occurs when ChatGPT generates nonsensical or false information (misinformation). &       & 157 \\
\cmidrule{1-1}\cmidrule{3-3}\cmidrule{5-5}\cmidrule{7-7}\cmidrule{9-9}\cmidrule{11-11}    209   &       & Frieder et al. \cite{frieder2023mathematical} &       & 2023  &       & Mathematical Setting &       & "AI hallucination" occurs when GPT, after answering correctly or incorrectly, tells the user unrelated information. &       & 126 \\
\cmidrule{1-1}\cmidrule{3-3}\cmidrule{5-5}\cmidrule{7-7}\cmidrule{9-9}\cmidrule{11-11}    210   &       & Borji \cite{borji2023categorical} &       & 2023  &       & ChatGPT &       & "AI hallucination" refers to inaccuracies in information or statements that are not in accordance with reality or the truth, often unintentional but resulting in incorrect or misleading information, particularly in the context of chatbots. Also called "Factual errors." &       & 158 \\
\cmidrule{1-1}\cmidrule{3-3}\cmidrule{5-5}\cmidrule{7-7}\cmidrule{9-9}\cmidrule{11-11}    211   &       & OpenAI \cite{openai2023gpt4} &       & 2023  &       & ChatGPT &       & "AI hallucination" occurs when ChatGPT produces content that is nonsensical or untruthful in relation to certain sources. &       & 0 \\
\cmidrule{1-1}\cmidrule{3-3}\cmidrule{5-5}\cmidrule{7-7}\cmidrule{9-9}\cmidrule{11-11}    212   &       & Manakul et al. \cite{manakul2023selfcheckgpt} &       & 2023  &       & Large Language Model &       & "AI hallucination" refers to the tendency of LLMs to hallucinate facts and fabricate information. &       & 48 \\
\cmidrule{1-1}\cmidrule{3-3}\cmidrule{5-5}\cmidrule{7-7}\cmidrule{9-9}\cmidrule{11-11}    213   &       & Nori et al. \cite{nori2023capabilities} &       & 2023  &       & Health &       & "AI hallucination" refers to erroneous generations by ChatGPT. &       & 127 \\
\cmidrule{1-1}\cmidrule{3-3}\cmidrule{5-5}\cmidrule{7-7}\cmidrule{9-9}\cmidrule{11-11}    214   &       & Li et al. \cite{li2023halueval} &       & 2023  &       & Large Language Model &       & "AI hallucination" occurs when LLMs generate content that conflicts with the source or cannot be verified by factual knowledge. &       & 4 \\
\cmidrule{1-1}\cmidrule{3-3}\cmidrule{5-5}\cmidrule{7-7}\cmidrule{9-9}\cmidrule{11-11}    215   &       & Zhao et al. \cite{zhao2023survey} &       & 2023  &       & Large Language Model &       & "Intrinsic hallucination" refers to the generated information in conflict with the existing source.\newline{}"Extrinsic hallucination" refers to the generated information that cannot be verified by the available source. &       & 188 \\
\cmidrule{1-1}\cmidrule{3-3}\cmidrule{5-5}\cmidrule{7-7}\cmidrule{9-9}\cmidrule{11-11}    216   &       & Adlakha et al. \cite{adlakha2023evaluating} &       & 2023  &       & Question Answering &       & "AI hallucination" occurs when conversational models produce factually incorrect or unsupported statements. &       & 5 \\
\cmidrule{1-1}\cmidrule{3-3}\cmidrule{5-5}\cmidrule{7-7}\cmidrule{9-9}\cmidrule{11-11}    217   &       & Athavale et al. \cite{athavale2023potential} &       & 2023  &       & Health &       & "AI hallucination" refers to generating syntactically correct but factually incorrect responses that seem plausible. &       & 0 \\
\cmidrule{1-1}\cmidrule{3-3}\cmidrule{5-5}\cmidrule{7-7}\cmidrule{9-9}\cmidrule{11-11}    218   &       & Chen et al. \cite{chen2023chat} &       & 2023  &       & Health &       & "AI hallucination" occurs when ChatGPT makes assumptions on details not provided in the input data. &       & 1 \\
\cmidrule{1-1}\cmidrule{3-3}\cmidrule{5-5}\cmidrule{7-7}\cmidrule{9-9}\cmidrule{11-11}    219   &       & Stahl and Eke \cite{stahl2024ethics} &       & 2023  &       & ChatGPT &       & "AI hallucination" refers to mistakes that ChatGPT makes when generating text that is semantically correct but factually incorrect or even nonsensical. &       & 1 \\
\cmidrule{1-1}\cmidrule{3-3}\cmidrule{5-5}\cmidrule{7-7}\cmidrule{9-9}\cmidrule{11-11}    220   &       & Walters and Wilder \cite{walters2023fabrication} &       & 2023  &       & Academia &       & "AI hallucination" occurs when ChatGPT provides factually incorrect responses. &       & 0 \\
\cmidrule{1-1}\cmidrule{3-3}\cmidrule{5-5}\cmidrule{7-7}\cmidrule{9-9}\cmidrule{11-11}    221   &       & Munro and Hope \cite{munro2023artificial} &       & 2023  &       & Health/ Academia &       & "AI hallucination" occurs when ChatGPT provided confident responses that seemed faithful and nonsensical. &       & 0 \\
\cmidrule{1-1}\cmidrule{3-3}\cmidrule{5-5}\cmidrule{7-7}\cmidrule{9-9}\cmidrule{11-11}    222   &       & Hashimoto and Johnson \cite{hashimoto2023use} &       & 2023  &       & Health/ Academia &       & "AI hallucination" occurs when LLM-generated text is based on the statistical associations of patterns of words to those seen in training data and prompts, resulting in elements such as unnecessary and unnatural repetition, lack of clarity or specificity, out-of-context content, inconsistent phrasing and arguments (particularly across long essays), incorrect word selection, and occasional incorrect statements presented as facts. &       & 1 \\
\cmidrule{1-1}\cmidrule{3-3}\cmidrule{5-5}\cmidrule{7-7}\cmidrule{9-9}\cmidrule{11-11}    223   &       & Jain \cite{jain2023chatgpt} &       & 2023  &       & ChatGPT &       & "AI hallucination" occurs when ChatGPT occasionally generates responses that appear correct but are nonsensical and discordant from real-world data, resulting in data inaccuracy. &       & 0 \\
\cmidrule{1-1}\cmidrule{3-3}\cmidrule{5-5}\cmidrule{7-7}\cmidrule{9-9}\cmidrule{11-11}    224   &       & Ho et al. \cite{ho2023chatgpt} &       & 2023  &       & ChatGPT &       & "AI hallucination" occurs when ChatGPT "hallucinates" false citations that appear convincingly real but cannot be located in any medical database. &       & 0 \\
\cmidrule{1-1}\cmidrule{3-3}\cmidrule{5-5}\cmidrule{7-7}\cmidrule{9-9}\cmidrule{11-11}    225   &       & Currie \cite{currie2023academic} &       & 2023  &       & ChatGPT &       & "AI hallucination" refer to false or misleading information produced by AI. A hallucination is a plausible response that is incorrect (it seems correct to ChatGPT but is not). &       & 21 \\
\cmidrule{1-1}\cmidrule{3-3}\cmidrule{5-5}\cmidrule{7-7}\cmidrule{9-9}\cmidrule{11-11}    226   &       & Liu et al. \cite{liu2023webglm} &       & 2023  &       & Question Answering &       & "AI hallucination" is LLM-generated answer which is not factual-grounded and sometimes severely wrong. &       & 1 \\
\cmidrule{1-1}\cmidrule{3-3}\cmidrule{5-5}\cmidrule{7-7}\cmidrule{9-9}\cmidrule{11-11}    227   &       & Campbell and Jovanovi{\'c} \cite{campbell2023detecting} &       & 2023  &       & Generative AI &       & "AI hallucination" refers to the propensity to generate assertions with no factual data, often deceives users into believing they are accurate. &       & 0 \\
\cmidrule{1-1}\cmidrule{3-3}\cmidrule{5-5}\cmidrule{7-7}\cmidrule{9-9}\cmidrule{11-11}    228   &       & Kshetri \cite{kshetri2023economics} &       & 2023  &       & Generative AI &       & "AI hallucination" refers to ChatGPT’s results that are incomplete or misleading. &       & 0 \\
\cmidrule{1-1}\cmidrule{3-3}\cmidrule{5-5}\cmidrule{7-7}\cmidrule{9-9}\cmidrule{11-11}    229   &       & Ali et al. \cite{ali2023addressing} &       & 2023  &       & Health &       & "AI hallucination" occurs when AI creates ungrounded, subtly incorrect information without self-awareness. &       & 1 \\
\cmidrule{1-1}\cmidrule{3-3}\cmidrule{5-5}\cmidrule{7-7}\cmidrule{9-9}\cmidrule{11-11}    230   &       & Tay \cite{tay2023chatgpt} &       & 2023  &       & Health/ Academia &       & "AI hallucination" refers to the phenomenon by which ChatGPT could convincingly produce factually inaccurate statements &       & 0 \\
\cmidrule{1-1}\cmidrule{3-3}\cmidrule{5-5}\cmidrule{7-7}\cmidrule{9-9}\cmidrule{11-11}    231   &       & Xu and Cohen \cite{xu2023lightweight} &       & 2023  &       & Text Summarization &       & "AI hallucination" occurs when the model confidently generates false information. &       & 1 \\
\cmidrule{1-1}\cmidrule{3-3}\cmidrule{5-5}\cmidrule{7-7}\cmidrule{9-9}\cmidrule{11-11}    232   &       & Kernan Ferier et al. \cite{kernan2023harnessing} &       & 2023  &       & Large Language Model &       & "AI hallucination" commonly refers to the phenomenon where LLMs occasionally generate outputs that, although appearing plausible, deviate from the user input, previously generated context, or factual knowledge.\newline{}\newline{}"Intrinsic hallucination" refers to contradictions between source material (e.g., training data and prompts), where generated content contradicts the information present in the source material.\newline{}\newline{}"Extrinsic hallucination" refers to information generated that cannot be verified by the source material, and may include content that lacks support or contradiction within the provided source data. &       & 0 \\
\cmidrule{1-1}\cmidrule{3-3}\cmidrule{5-5}\cmidrule{7-7}\cmidrule{9-9}\cmidrule{11-11}    233   &       & Tsai et al. \cite{tsai2023exploring} &       & 2023  &       & Legal Setting &       & "AI hallucination" occurs when LLMs generate misleading text and information. &       & 3 \\
\cmidrule{1-1}\cmidrule{3-3}\cmidrule{5-5}\cmidrule{7-7}\cmidrule{9-9}\cmidrule{11-11}    234   &       & Xin et al. \cite{xin2023dialogue} &       & 2023  &       & Dialogue &       & "AI hallucination" refers to the occurrence where pre-trained text generation models occasionally generate text that is nonsensical or unfaithful to the provided source input. &       & 0 \\
\cmidrule{1-1}\cmidrule{3-3}\cmidrule{5-5}\cmidrule{7-7}\cmidrule{9-9}\cmidrule{11-11}    235   &       & Murgia et al. \cite{murgia2023chatgpt} &       & 2023  &       & ChatGPT &       & "AI hallucination" occurs when generative LLMs generate unintended text, leading to degraded system performance and unmet user expectations in real-world scenarios. &       & 1 \\
\cmidrule{1-1}\cmidrule{3-3}\cmidrule{5-5}\cmidrule{7-7}\cmidrule{9-9}\cmidrule{11-11}    236   &       & Kunze et al. \cite{kunze2023s} &       & 2023  &       & Large Language Model &       & "AI hallucination" refers to the phenomenon where the GPT model appears to be designed to provide incorrect answers rather than admit this to its users. &       & 0 \\
\cmidrule{1-1}\cmidrule{3-3}\cmidrule{5-5}\cmidrule{7-7}\cmidrule{9-9}\cmidrule{11-11}    237   &       & Cascella \cite{cascella2023writing} &       & 2023  &       & Academia &       & "AI hallucination" refers to ChatGPT generating answers that sound credible but may be incorrect or nonsensical. &       & 0 \\
\cmidrule{1-1}\cmidrule{3-3}\cmidrule{5-5}\cmidrule{7-7}\cmidrule{9-9}\cmidrule{11-11}    238   &       & Watanabe \cite{watanabe2023exploring} &       & 2023  &       & Government &       & "AI hallucination" refers to the phenomenon where AI text generators produce statements that are consistent with the system's internal logic but are not based on any true context or source. &       & 0 \\
\cmidrule{1-1}\cmidrule{3-3}\cmidrule{5-5}\cmidrule{7-7}\cmidrule{9-9}\cmidrule{11-11}    239   &       & Nakaura et al. \cite{nakaura2023preliminary} &       & 2023  &       & Health &       & "AI hallucination" occurs when LLMs like GPT series have the potential to generate inaccurate content. &       & 0 \\
\cmidrule{1-1}\cmidrule{3-3}\cmidrule{5-5}\cmidrule{7-7}\cmidrule{9-9}\cmidrule{11-11}    240   &       & Feuerriegel et al. \cite{feuerriegel2023generative} &       & 2023  &       & Generative AI &       & "AI hallucination" refers to mistakes in the generated text that are semantically or syntactically plausible but are actually nonsensical or incorrect. In other words, the generative AI model produces content that is not based on any facts or evidence, but rather on its own assumptions or biases. Moreover, the output of generative AI, especially that of LLMs, is typically not easily verifiable. &       & 0 \\
\cmidrule{1-1}\cmidrule{3-3}\cmidrule{5-5}\cmidrule{7-7}\cmidrule{9-9}\cmidrule{11-11}    241   &       & Hryciw et al. \cite{hryciw2023doctor} &       & 2023  &       & Health &       & "AI hallucination" occurs when AI algorithms, which are not infallible, produce false or misleading information. &       & 0 \\
\cmidrule{1-1}\cmidrule{3-3}\cmidrule{5-5}\cmidrule{7-7}\cmidrule{9-9}\cmidrule{11-11}    242   &       & Bran et al. \cite{bran2023emerging} &       & 2023  &       & Generative AI &       & "AI hallucination" refers to nonsense expressed in an authoritative tone and filling knowledge gaps with falsehoods. &       & 5 \\
\cmidrule{1-1}\cmidrule{3-3}\cmidrule{5-5}\cmidrule{7-7}\cmidrule{9-9}\cmidrule{11-11}    243   &       & Sovrano et al. \cite{sovrano2023toward} &       & 2023  &       & ChatGPT &       & "AI hallucination" occurs when content is generated that does not maintain fidelity to a given context or source. &       & 0 \\
\cmidrule{1-1}\cmidrule{3-3}\cmidrule{5-5}\cmidrule{7-7}\cmidrule{9-9}\cmidrule{11-11}    244   &       & Mishra et al. \cite{mishra2023tpack} &       & 2023  &       & ChatGPT &       & "AI hallucination" occurs when models produce outputs that are not grounded in their training data. &       & 0 \\
\cmidrule{1-1}\cmidrule{3-3}\cmidrule{5-5}\cmidrule{7-7}\cmidrule{9-9}\cmidrule{11-11}    245   &       & Lam et al. \cite{lam2023applying} &       & 2023  &       & Legal Setting &       & "AI hallucination" refers to LLMs output results are not realistic, do not follow user given context or match any data patterns that it has been trained on. &       & 1 \\
\cmidrule{1-1}\cmidrule{3-3}\cmidrule{5-5}\cmidrule{7-7}\cmidrule{9-9}\cmidrule{11-11}    246   &       & Wan et al. \cite{wan2023faithfulness} &       & 2023  &       & Text Summarization &       & "AI hallucination" occurs when the generated summary contains facts or entities not present in the original document. &       & 4 \\
\cmidrule{1-1}\cmidrule{3-3}\cmidrule{5-5}\cmidrule{7-7}\cmidrule{9-9}\cmidrule{11-11}    247   &       & Houston and Corrado \cite{houston2023embracing} &       & 2023  &       & Academia &       & "AI hallucination" commonly refers to the phenomenon where LLMs occasionally generate outputs that, although appearing plausible, deviate from the user input, previously generated context, or factual knowledge. &       & 2 \\
\cmidrule{1-1}\cmidrule{3-3}\cmidrule{5-5}\cmidrule{7-7}\cmidrule{9-9}\cmidrule{11-11}    248   &       & Gonz{\'a}lez-Mora et al. \cite{gonzalez2023improving} &       & 2023  &       & Natural Language Generation &       & "AI hallucination" refers to the generation of texts that are apparently well-written but unsubstantiated and not faithful to the provided data. &       & 2 \\
\cmidrule{1-1}\cmidrule{3-3}\cmidrule{5-5}\cmidrule{7-7}\cmidrule{9-9}\cmidrule{11-11}    249   &       & Lim et al. \cite{lim2023you} &       & 2023  &       & Dialogue &       & "AI hallucination" refers to situations where the generated output contradicts the reference knowledge and includes instances when the generated output cannot be confirmed from the knowledge source. &       & 0 \\
\cmidrule{1-1}\cmidrule{3-3}\cmidrule{5-5}\cmidrule{7-7}\cmidrule{9-9}\cmidrule{11-11}    250   &       & Alowais et al. \cite{alowais2023revolutionizing} &       & 2023  &       & Health &       & "AI hallucination" refers to the tendency of AI-generated data and/or analysis to fabricate and create false information that cannot be supported by existing evidence, even though it may appear realistic and convincing. &       & 2 \\
\cmidrule{1-1}\cmidrule{3-3}\cmidrule{5-5}\cmidrule{7-7}\cmidrule{9-9}\cmidrule{11-11}    251   &       & Liu et al. \cite{liu2023using} &       & 2023  &       & Health &       & "AI hallucination" can be defined as the capability of LLM-based chatbots to convey false information as though it were true. &       & 33 \\
\cmidrule{1-1}\cmidrule{3-3}\cmidrule{5-5}\cmidrule{7-7}\cmidrule{9-9}\cmidrule{11-11}    252   &       & Xie et al. \cite{xie2023faithful} &       & 2023  &       & Health &       & "AI hallucination" refers to when a medical AI system is considered unfaithful or to have a factual inconsistency issue, as it generates content that is not supported by existing knowledge, reference, or data.\newline{}\newline{}Intrinsic Error: The generated output contradicts with existing knowledge, reference, or data. \newline{}\newline{}Extrinsic Error: The generated output cannot be confirmed (either supported or contradicted) by existing knowledge, reference, or data.  &       & 0 \\
\cmidrule{1-1}\cmidrule{3-3}\cmidrule{5-5}\cmidrule{7-7}\cmidrule{9-9}\cmidrule{11-11}    253   &       & Bernstein et al. \cite{bernstein2023comparison} &       & 2023  &       & Health &       & "AI hallucination" refers to chatbot outputs that sound convincingly plausible yet are factually inaccurate. &       & 0 \\
\cmidrule{1-1}\cmidrule{3-3}\cmidrule{5-5}\cmidrule{7-7}\cmidrule{9-9}\cmidrule{11-11}    254   &       & Heck \cite{heck2023artificial} &       & 2023  &       & Health &       & "AI hallucination" refers to ChatGPT providing text that combines real and fabricated evidence, resulting in plausible answers, occasionally appearing as nonsense when assessed according to the common knowledge of experts in different areas, also called "mistakes."\newline{}\newline{}“Intrinsic hallucinations” occur when the generation output contradicts the source content.\newline{}\newline{}“Extrinsic hallucinations” occur when the output can neither be supported nor contradicted by the source. &       & 3 \\
\cmidrule{1-1}\cmidrule{3-3}\cmidrule{5-5}\cmidrule{7-7}\cmidrule{9-9}\cmidrule{11-11}    255   &       & Ray and Majumder \cite{ray2023potential} &       & 2023  &       & Health &       & "AI hallucination" refers to the generation of inaccurate or false information by AI-based systems such as ChatGPT. &       & 0 \\
\cmidrule{1-1}\cmidrule{3-3}\cmidrule{5-5}\cmidrule{7-7}\cmidrule{9-9}\cmidrule{11-11}    256   &       & Ang et al. \cite{ang2023rise} &       & 2023  &       & Health/ Academia &       & "AI hallucination" refers to the phenomenon where text generated by ChatGPT may appear credible but can be pure confabulation, containing a combination of both facts and fabricated information, or entirely fictitious pseudoscientific material. &       & 5 \\
\cmidrule{1-1}\cmidrule{3-3}\cmidrule{5-5}\cmidrule{7-7}\cmidrule{9-9}\cmidrule{11-11}    257   &       & Talyshinskii et al. \cite{talyshinskii2023expanding} &       & 2023  &       & Health/ Academia &       & "AI hallucination" refers to a phenomenon in writing influenced more by learned patterns than scientific facts, which leads to mistakes. &       & 0 \\
\cmidrule{1-1}\cmidrule{3-3}\cmidrule{5-5}\cmidrule{7-7}\cmidrule{9-9}\cmidrule{11-11}    258   &       & Kung et al. \cite{kung2023evaluating} &       & 2023  &       & Health &       & "AI hallucination" refers to the phenomenon where ChatGPT, while citing a verifiable source, may draw information that is outdated or entirely incorrect, despite providing logical justifications for its answer choices, thus leading to logical errors and assertions of false facts. &       & 0 \\
\cmidrule{1-1}\cmidrule{3-3}\cmidrule{5-5}\cmidrule{7-7}\cmidrule{9-9}\cmidrule{11-11}    259   &       & Friederichs et al. \cite{friederichs2023chatgpt} &       & 2023  &       & Health/ Academia &       & "AI hallucination" refers to the behavior in which wrong answers are just as convincingly justified as correct ones, a phenomenon not uncommon in large language models. &       & 5 \\
\cmidrule{1-1}\cmidrule{3-3}\cmidrule{5-5}\cmidrule{7-7}\cmidrule{9-9}\cmidrule{11-11}    260   &       & Ghorashi et al. \cite{ghorashi2023ai} &       & 2023  &       & Health/ Academia &       & "AI hallucination" occurs when chatbots have the potential to falsify and create references. &       & 0 \\
\cmidrule{1-1}\cmidrule{3-3}\cmidrule{5-5}\cmidrule{7-7}\cmidrule{9-9}\cmidrule{11-11}    261   &       & McBee et al. \cite{mcbee2023interdisciplinary} &       & 2023  &       & Sport &       & "AI hallucination" occurs as the generation of unsupported or false information, which is a prevalent issue with LLM-based chatbots. &       & 0 \\
\cmidrule{1-1}\cmidrule{3-3}\cmidrule{5-5}\cmidrule{7-7}\cmidrule{9-9}\cmidrule{11-11}    262   &       & Reis \cite{reis2023chatgpt} &       & 2023  &       & Health &       & "AI hallucination" may generate plausible sounding but incorrect or nonsensical answers that does not seem to be justified by its training data, such as claim to be human. &       & 1 \\
\cmidrule{1-1}\cmidrule{3-3}\cmidrule{5-5}\cmidrule{7-7}\cmidrule{9-9}\cmidrule{11-11}    263   &       & Joachimiak et al. \cite{joachimiak2023gene} &       & 2023  &       & Genetic &       & "AI hallucination" ocuurse when LLM model fabricated a term for a gene set. &       & 0 \\
\cmidrule{1-1}\cmidrule{3-3}\cmidrule{5-5}\cmidrule{7-7}\cmidrule{9-9}\cmidrule{11-11}    264   &       & Chen \cite{chen2023research} &       & 2023  &       & Academia &       & "AI hallucination" refers to false or nonsense information presented as fact by an LLM. &       & 0 \\
\cmidrule{1-1}\cmidrule{3-3}\cmidrule{5-5}\cmidrule{7-7}\cmidrule{9-9}\cmidrule{11-11}    265   &       & Delsoz et al. \cite{delsoz2023performance} &       & 2023  &       & Health/ ChatGPT &       & "AI hallucination" occurs when ChatGPT generates responses that appear fluent and believable but may contain factual inaccuracies. &       & 0 \\
\cmidrule{1-1}\cmidrule{3-3}\cmidrule{5-5}\cmidrule{7-7}\cmidrule{9-9}\cmidrule{11-11}    266   &       & Takagi et al. \cite{takagi2023performance} &       & 2023  &       & Health/ ChatGPT &       & "AI hallucination" is defined as producing nonsensical or untruthful content concerning certain sources. &       & 11 \\
\cmidrule{1-1}\cmidrule{3-3}\cmidrule{5-5}\cmidrule{7-7}\cmidrule{9-9}\cmidrule{11-11}    267   &       & Kaneda et al. \cite{kaneda2023assessing} &       & 2023  &       & Health/ ChatGPT &       & "AI hallucination" occurs when ChatGPT provides erroneous information in a naturalistic manner. &       & 3 \\
\cmidrule{1-1}\cmidrule{3-3}\cmidrule{5-5}\cmidrule{7-7}\cmidrule{9-9}\cmidrule{11-11}    268   &       & Ahn \cite{ahn2023impending} &       & 2023  &       & Health &       & "AI hallucination" refers to the generation of false or logically incorrect text that appears plausible and grammatically correct. &       & 12 \\
\cmidrule{1-1}\cmidrule{3-3}\cmidrule{5-5}\cmidrule{7-7}\cmidrule{9-9}\cmidrule{11-11}    269   &       & Jin et al. \cite{jin2023matching} &       & 2023  &       & Health &       & "AI hallucination" refers to the occurrence where LLMs usually produce plausible-sounding but incorrect outputs. &       & 0 \\
\cmidrule{1-1}\cmidrule{3-3}\cmidrule{5-5}\cmidrule{7-7}\cmidrule{9-9}\cmidrule{11-11}    270   &       & Ahn \cite{ahn2023use} &       & 2023  &       & Health/ ChatGPT &       & "AI hallucination" refers to a phenomenon where outputs may deviate from factual accuracy or provided context. &       & 0 \\
\cmidrule{1-1}\cmidrule{3-3}\cmidrule{5-5}\cmidrule{7-7}\cmidrule{9-9}\cmidrule{11-11}    271   &       & Liu et al. \cite{liu2023utility} &       & 2023  &       & Health/ ChatGPT &       & “AI hallucination” refers to the fact that the content generated by the model is not based on reality, creating a completely fabricated story or fact. &       & 20 \\
\cmidrule{1-1}\cmidrule{3-3}\cmidrule{5-5}\cmidrule{7-7}\cmidrule{9-9}\cmidrule{11-11}    272   &       & Doorn \cite{doorn2023use} &       & 2023  &       & Water Domain &       & "AI hallucination" refers to LLMs' tendency to generate factually nonsensical or incorrect text. &       & 0 \\
\cmidrule{1-1}\cmidrule{3-3}\cmidrule{5-5}\cmidrule{7-7}\cmidrule{9-9}\cmidrule{11-11}    273   &       & Hiesinger et al. \cite{hiesinger2023almanac} &       & 2023  &       & Health &       & "AI hallucination" refers to the propensity of LLMs to generate factually incorrect statements. &       & 3 \\
\cmidrule{1-1}\cmidrule{3-3}\cmidrule{5-5}\cmidrule{7-7}\cmidrule{9-9}\cmidrule{11-11}    274   &       & Palal et al. \cite{palal2023chatgpt} &       & 2023  &       & Health/ ChatGPT &       & "AI hallucination" occurs when ChatGPT generates inaccurate or contradictory information. &       & 0 \\
\cmidrule{1-1}\cmidrule{3-3}\cmidrule{5-5}\cmidrule{7-7}\cmidrule{9-9}\cmidrule{11-11}    275   &       & Coskun et al. \cite{coskun2023assessing} &       & 2023  &       & Question Answering/ ChatGPT &       & "AI hallucination" refers to an instance where the model generates information that is not supported by existing evidence or factual data. &       & 0 \\
\cmidrule{1-1}\cmidrule{3-3}\cmidrule{5-5}\cmidrule{7-7}\cmidrule{9-9}\cmidrule{11-11}    276   &       & Baldassarre et al. \cite{baldassarre2023social} &       & 2023  &       & ChatGPT &       & "AI hallucination" refers to the phenomenon of inaccurate information. &       & 0 \\
\cmidrule{1-1}\cmidrule{3-3}\cmidrule{5-5}\cmidrule{7-7}\cmidrule{9-9}\cmidrule{11-11}    277   &       & Boujemaa et al. \cite{boujemaa2023responsible} &       & 2023  &       & Retail &       & "AI hallucination" occures where the model produces untruthful or misleading information. &       & 0 \\
\cmidrule{1-1}\cmidrule{3-3}\cmidrule{5-5}\cmidrule{7-7}\cmidrule{9-9}\cmidrule{11-11}    278   &       & Li et al. \cite{li2023improving} &       & 2023  &       & Dialogue &       & "AI hallucination" are NLP generated content that appear to be relevant bbut are not actually faithful to the underlting text. &       & 0 \\
\cmidrule{1-1}\cmidrule{3-3}\cmidrule{5-5}\cmidrule{7-7}\cmidrule{9-9}\cmidrule{11-11}    279   &       & Urban et al. \cite{urban2023omniscientdb} &       & 2023  &       & Database &       & "AI hallucination" is a phenomenon where LLMs generate non-factual statements. &       & 1 \\
\cmidrule{1-1}\cmidrule{3-3}\cmidrule{5-5}\cmidrule{7-7}\cmidrule{9-9}\cmidrule{11-11}    280   &       & Mahon et al. \cite{10.1145/3610969.3610982} &       & 2023  &       & Academia &       & "AI hallucination" is the generation of output that appears convincing but is factually untrue or unrelated to the current context. &       & 0 \\
\cmidrule{1-1}\cmidrule{3-3}\cmidrule{5-5}\cmidrule{7-7}\cmidrule{9-9}\cmidrule{11-11}    281   &       & August et al. \cite{august2022paper} &       & 2023  &       & Health &       & "AI hallucination" refers to the limitation in current text generation capabilities, which carries the risk of generating factually incorrect or inconsistent text. &       & 20 \\
\cmidrule{1-1}\cmidrule{3-3}\cmidrule{5-5}\cmidrule{7-7}\cmidrule{9-9}\cmidrule{11-11}    282   &       & Fischer \cite{fischer2023generative} &       & 2023  &       & Legal Setting &       & "AI hallucination" refers to a phenomenon where an LM, or Language Model, is a system for haphazardly stitching together sequences of linguistic forms it has observed in its vast training data, according to probabilistic information about how they combine, but without any reference to meaning. Also called "Stochastic Parrots." &       & 0 \\
\cmidrule{1-1}\cmidrule{3-3}\cmidrule{5-5}\cmidrule{7-7}\cmidrule{9-9}\cmidrule{11-11}    283   &       & Lee et al. \cite{lee2023dapie} &       & 2023  &       & Dialogue &       & "AI hallucination" are LLMs generated information that is non-factual or nonsensical. &       & 3 \\
\cmidrule{1-1}\cmidrule{3-3}\cmidrule{5-5}\cmidrule{7-7}\cmidrule{9-9}\cmidrule{11-11}    284   &       & Scotti et al. \cite{scotti2023primer} &       & 2023  &       & Chatbot &       & "AI hallucination" refers to the case where agents generate responses without actually knowing the information it is talking about or without referring to some knowledge base, leading to possibly wrong/misleading information. &       & 0 \\
\cmidrule{1-1}\cmidrule{3-3}\cmidrule{5-5}\cmidrule{7-7}\cmidrule{9-9}\cmidrule{11-11}    285   &       & Zhan et al. \cite{zhan2023deceptive} &       & 2023  &       & ChatGPT &       & "AI hallucination" refers to the phenomenon, also known as the "hallucination effect," where chatbots like ChatGPT may generate misleading and deceptive information that can have adverse impacts on users who may struggle to distinguish fact from fiction. &       & 0 \\
\cmidrule{1-1}\cmidrule{3-3}\cmidrule{5-5}\cmidrule{7-7}\cmidrule{9-9}\cmidrule{11-11}    286   &       & Vargas-Murillo et al. \cite{vargas2023ethics} &       & 2023  &       & Academia &       & "AI hallucination" refers to the phenomenon, also known as the "hallucination effect," which causes an AI to invent familiar terms. &       & 0 \\
\cmidrule{1-1}\cmidrule{3-3}\cmidrule{5-5}\cmidrule{7-7}\cmidrule{9-9}\cmidrule{11-11}    287   &       & White et al. \cite{white2023clinidigest} &       & 2023  &       & Text Summarization &       & "AI hallucination" refers to the tendency to produce content that is nonsensical or untruthful in relation to certain sources. &       & 0 \\
\cmidrule{1-1}\cmidrule{3-3}\cmidrule{5-5}\cmidrule{7-7}\cmidrule{9-9}\cmidrule{11-11}    288   &       & Cusumano \cite{cusumano2023generative} &       & 2023  &       & Technology &       & "AI hallucination" occurs when LLMs, in the absence of an answer to a query, use predictive analytics to make up reasonable but sometimes incorrect responses. &       & 0 \\
\cmidrule{1-1}\cmidrule{3-3}\cmidrule{5-5}\cmidrule{7-7}\cmidrule{9-9}\cmidrule{11-11}    289   &       & Nashid et al. \cite{nashid2023retrieval} &       & 2023  &       & Large Language Model &       & "AI hallucination" refers to ChatGPT's outputs that are incorrect or nonsensical. &       & 12 \\
\cmidrule{1-1}\cmidrule{3-3}\cmidrule{5-5}\cmidrule{7-7}\cmidrule{9-9}\cmidrule{11-11}    290   &       & Pataranutaporn et al. \cite{pataranutaporn2023living} &       & 2023  &       & Technology &       & "AI hallucination" refers to the phenomenon where ChatGPT generates content that is nonsensical or unfaithful to the provided source content. &       & 2 \\
\cmidrule{1-1}\cmidrule{3-3}\cmidrule{5-5}\cmidrule{7-7}\cmidrule{9-9}\cmidrule{11-11}    291   &       & Faggioli et al. \cite{faggioli2023perspectives} &       & 2023  &       & Large Language Model &       & "AI hallucination" occurs when LLMs generate text that contains inaccurate or false information, often in an affirmative manner that makes it difficult for humans to even suspect errors. &       & 0 \\
\cmidrule{1-1}\cmidrule{3-3}\cmidrule{5-5}\cmidrule{7-7}\cmidrule{9-9}\cmidrule{11-11}    292   &       & Marczak-Czajka and Cleland-Huang \cite{10260859} &       & 2023  &       & Human-Value Story &       & "AI hallucination" occurs when generative models produce output that is nonsensical with incorrect, bizarre logic. &       & 0 \\
\cmidrule{1-1}\cmidrule{3-3}\cmidrule{5-5}\cmidrule{7-7}\cmidrule{9-9}\cmidrule{11-11}    293   &       & Stefńnska et al. \cite{10261132} &       & 2023  &       & Quantum Physics &       & "AI hallucination" occurs when LLMs generate data not observed in the training dataset. &       & 0 \\
\cmidrule{1-1}\cmidrule{3-3}\cmidrule{5-5}\cmidrule{7-7}\cmidrule{9-9}\cmidrule{11-11}    294   &       & Ezenkwu \cite{10248647} &       & 2023  &       & Customer Service &       & "AI hallucination" occurs when ChatGPT generates responses that are irrelevant or incorrect. &       & 0 \\
\cmidrule{1-1}\cmidrule{3-3}\cmidrule{5-5}\cmidrule{7-7}\cmidrule{9-9}\cmidrule{11-11}    295   &       & Fayyad \cite{10148666} &       & 2023  &       & ChatGPT &       & "AI hallucination" occurs when generative AI models lose track of the source of information, lacking reasoning capability or semantic understanding, and instead resort to autocompletion through pattern matching, resulting in making things up. &       & 0 \\
\cmidrule{1-1}\cmidrule{3-3}\cmidrule{5-5}\cmidrule{7-7}\cmidrule{9-9}\cmidrule{11-11}    296   &       & Mrabet and Studholme \cite{mrabet2023chatgpt} &       & 2023  &       & ChatGPT &       & "AI hallucination" refers to The AI’s inability to understand what it has written is clear. &       & 2 \\
\cmidrule{1-1}\cmidrule{3-3}\cmidrule{5-5}\cmidrule{7-7}\cmidrule{9-9}\cmidrule{11-11}    297   &       & Pitt \cite{pitt2023chatsh} &       & 2023  &       & Academia &       & "AI hallucination" occurs when the AI/LLM produces a plausible output that, however, does not seem to be warranted by the training data. &       & 0 \\
\cmidrule{1-1}\cmidrule{3-3}\cmidrule{5-5}\cmidrule{7-7}\cmidrule{9-9}\cmidrule{11-11}    298   &       & Crosthwaite and Baisa \cite{crosthwaite2023generative} &       & 2023  &       & ChatGPT &       & "AI hallucination" occurs when ChatGPT invents terms that lie outside of its training data. &       & 1 \\
\cmidrule{1-1}\cmidrule{3-3}\cmidrule{5-5}\cmidrule{7-7}\cmidrule{9-9}\cmidrule{11-11}    299   &       & Vinny \cite{vinny2023invoking} &       & 2023  &       & Health &       & "AI hallucination" occurs when LLMs generate erroneous medical information to support their opinions &       & 0 \\
\cmidrule{1-1}\cmidrule{3-3}\cmidrule{5-5}\cmidrule{7-7}\cmidrule{9-9}\cmidrule{11-11}    300   &       & Solyman et al. \cite{solyman2023optimizing} &       & 2023  &       & Grammar &       & "AI hallucination" occurs when the system produces translations that are completely inadequate due to an overreliance on the target context in NMT. &       & 4 \\
\cmidrule{1-1}\cmidrule{3-3}\cmidrule{5-5}\cmidrule{7-7}\cmidrule{9-9}\cmidrule{11-11}    301   &       & Waqas et al. \cite{waqas2023revolutionizing} &       & 2023  &       & Health &       & "AI hallucination" refers to a known limitation of generative AI, encompassing mistakes in the generated text or images that are semantically, syntactically, or visually plausible but are, in fact, incorrect, nonsensical, and do not refer to any real-world concepts. &       & 0 \\
\cmidrule{1-1}\cmidrule{3-3}\cmidrule{5-5}\cmidrule{7-7}\cmidrule{9-9}\cmidrule{11-11}    302   &       & Stephens et al. \cite{stephens2023battle} &       & 2023  &       & Health &       & "AI hallucination" occurs when chatbots may propagate erroneous information or even make up information. &       & 0 \\
\cmidrule{1-1}\cmidrule{3-3}\cmidrule{5-5}\cmidrule{7-7}\cmidrule{9-9}\cmidrule{11-11}    303   &       & Dien \cite{dien2023generative} &       & 2023  &       & Health/ Academia &       & "AI hallucination" refers to the highly susceptible nature of ChatGPT to produce erroneous outputs. &       & 5 \\
\cmidrule{1-1}\cmidrule{3-3}\cmidrule{5-5}\cmidrule{7-7}\cmidrule{9-9}\cmidrule{11-11}    304   &       & Abu-Farha et al. \cite{abu2023evaluation} &       & 2023  &       & Health &       & "AI hallucination" refers to the generation of scientifically false content that might seem convincing to nonexperts. &       & 0 \\
\cmidrule{1-1}\cmidrule{3-3}\cmidrule{5-5}\cmidrule{7-7}\cmidrule{9-9}\cmidrule{11-11}    305   &       & Tippareddy et al. \cite{tippareddy2023radiology} &       & 2023  &       & Health/ Academia &       & "AI hallucination" refers to confident responses generated by ChatGPT without being justified by training data. &       & 0 \\
\cmidrule{1-1}\cmidrule{3-3}\cmidrule{5-5}\cmidrule{7-7}\cmidrule{9-9}\cmidrule{11-11}    306   &       & Oviedo-Trespalacios et al. \cite{oviedo2023risks} &       & 2023  &       & ChatGPT &       & "AI hallucination" refers to the occurrence where ChatGPT can produce answers that appear credible but may be incorrect or nonsensical. &       & 18 \\
\cmidrule{1-1}\cmidrule{3-3}\cmidrule{5-5}\cmidrule{7-7}\cmidrule{9-9}\cmidrule{11-11}    307   &       & Sarraju et al. \cite{sarraju2023opportunities} &       & 2023  &       & Health/ ChatGPT &       & "AI hallucination" refers to Inaccurate information may be presented in a confident manner, including nonexistent references to scientific literature. Also called "confabulation." &       & 0 \\
\cmidrule{1-1}\cmidrule{3-3}\cmidrule{5-5}\cmidrule{7-7}\cmidrule{9-9}\cmidrule{11-11}    308   &       & Coskun et al. \cite{COSKUN2023} &       & 2023  &       & Health/ ChatGPT &       & "AI hallucination" refers to situations where the AI generates responses that are not derived from its training data, potentially leading to inaccuracies or misunderstandings. &       & 0 \\
\cmidrule{1-1}\cmidrule{3-3}\cmidrule{5-5}\cmidrule{7-7}\cmidrule{9-9}\cmidrule{11-11}    309   &       & Lam \cite{lam2023chatgpt} &       & 2023  &       & Health &       & "AI hallucination" refers to occurrences where AI models output plausible but factually incorrect information. &       & 0 \\
\cmidrule{1-1}\cmidrule{3-3}\cmidrule{5-5}\cmidrule{7-7}\cmidrule{9-9}\cmidrule{11-11}    310   &       & Pantanowitz and Pantanowitz \cite{pantanowitz2023implications} &       & 2023  &       & Health/ ChatGPT &       & "AI hallucination" occurs when the AI imagines facts that are not real as being real or makes reasoning errors that it should not be making. &       & 0 \\
\cmidrule{1-1}\cmidrule{3-3}\cmidrule{5-5}\cmidrule{7-7}\cmidrule{9-9}\cmidrule{11-11}    311   &       & Boussen et al. \cite{boussen2023chatgpt} &       & 2023  &       & Health/ Academia &       & "AI hallucination" refers to the occurrence where a model might 'invent' information that seems plausible based on the patterns and structures it has learned. &       & 0 \\
\cmidrule{1-1}\cmidrule{3-3}\cmidrule{5-5}\cmidrule{7-7}\cmidrule{9-9}\cmidrule{11-11}    312   &       & Schlam et al. \cite{schlam2023artificial} &       & 2023  &       & Health/ Academia &       & "AI hallucination" refers to the phenomenon where NLPs often make subtle mistakes. &       & 0 \\
\cmidrule{1-1}\cmidrule{3-3}\cmidrule{5-5}\cmidrule{7-7}\cmidrule{9-9}\cmidrule{11-11}    313   &       & Lyu and Wu \cite{LYU20231} &       & 2023  &       & Academia &       & "AI hallucination" occurs when LLM responses can sometimes be repetitive or contain non-factual information. &       & 0 \\
\cmidrule{1-1}\cmidrule{3-3}\cmidrule{5-5}\cmidrule{7-7}\cmidrule{9-9}\cmidrule{11-11}    314   &       & Sparkes \cite{sparkes2023chatbots} &       & 2023  &       & Chatbot &       & "AI hallucination" refers to the phenomenon where an AI, in response to prompts, will produce convincing statements that are actually inaccurate or totally false &       & 0 \\
\cmidrule{1-1}\cmidrule{3-3}\cmidrule{5-5}\cmidrule{7-7}\cmidrule{9-9}\cmidrule{11-11}    315   &       & Bhatia and Kulkarni \cite{BHATIA2023103564} &       & 2023  &       & Academia &       & "AI hallucination" occurs when the AI sometimes writes plausible-sounding but incorrect, nonsensical answers. &       & 0 \\
\cmidrule{1-1}\cmidrule{3-3}\cmidrule{5-5}\cmidrule{7-7}\cmidrule{9-9}\cmidrule{11-11}    316   &       & Chatelan et al. \cite{chatelan2023chatgpt} &       & 2023  &       & Health &       & "AI hallucination" occurs when ChatGPT makes up or distorts facts, including the creation of made-up references. &       & 0 \\
\cmidrule{1-1}\cmidrule{3-3}\cmidrule{5-5}\cmidrule{7-7}\cmidrule{9-9}\cmidrule{11-11}    317   &       & Lareyre et al. \cite{lareyre2023comprehensive} &       & 2023  &       & Health &       & "AI hallucination" refers to the phenomenon where the model can generate content output that is incorrect or nonsensical, despite appearing reliable. &       & 0 \\
\cmidrule{1-1}\cmidrule{3-3}\cmidrule{5-5}\cmidrule{7-7}\cmidrule{9-9}\cmidrule{11-11}    318   &       & Wilkins \cite{wilkins2023robot} &       & 2023  &       & Health &       & "AI hallucination" occurs when the system erroneously generates “fantastical, unfaithful, or nonsensical outputs.” &       & 0 \\
\cmidrule{1-1}\cmidrule{3-3}\cmidrule{5-5}\cmidrule{7-7}\cmidrule{9-9}\cmidrule{11-11}    319   &       & Piazza et al. \cite{piazza2023long} &       & 2023  &       & Academia &       & "AI hallucination" refers to the phenomenon where ChatGPT may generate output that is grammatically correct and coherent but may not be appropriate for the intended audience or purpose. &       & 0 \\
\cmidrule{1-1}\cmidrule{3-3}\cmidrule{5-5}\cmidrule{7-7}\cmidrule{9-9}\cmidrule{11-11}    320   &       & Scanlon et al. \cite{scanlon2023digital} &       & 2023  &       & Academia &       & "AI hallucination" refers to the phenomenon where ChatGPT prioritize generating humanlike text in response to a prompt, often leading them to focus on providing an answer rather than the correct one, resulting in inaccurate or incorrect responses presented to users with an unfounded confidence, and even generating fake bibliographic information when asked for references, which could be misleading if not critically examined. &       & 2 \\
\cmidrule{1-1}\cmidrule{3-3}\cmidrule{5-5}\cmidrule{7-7}\cmidrule{9-9}\cmidrule{11-11}    321   &       & Kaneda \cite{kaneda2023chatgpt} &       & 2023  &       & Health &       & "AI hallucination" refers to a phenomenon where ChatGPT generates plausible but untrue responses. &       & 1 \\
\cmidrule{1-1}\cmidrule{3-3}\cmidrule{5-5}\cmidrule{7-7}\cmidrule{9-9}\cmidrule{11-11}    322   &       & Tan et al. \cite{tan2023generative} &       & 2023  &       & Health &       & "AI hallucination" refers to the occurrence where invented, inaccurate statements are presented as lucidly as accurate information. Also called "fact fabrication." &       & 0 \\
\cmidrule{1-1}\cmidrule{3-3}\cmidrule{5-5}\cmidrule{7-7}\cmidrule{9-9}\cmidrule{11-11}    323   &       & Ai et al. \cite{ai2023information} &       & 2023  &       & Information Retrieval &       & "AI hallucination" occurs when LLMs may occasionally generate erroneous or nonsensical responses. &       & 3 \\
\cmidrule{1-1}\cmidrule{3-3}\cmidrule{5-5}\cmidrule{7-7}\cmidrule{9-9}\cmidrule{11-11}    324   &       & Ruma et al. \cite{ruma2023transformer} &       & 2023  &       & Natural Language Generation &       & "AI hallucination" occurs when an NLG system generates unfaithful or nonfactual content. &       & 0 \\
\cmidrule{1-1}\cmidrule{3-3}\cmidrule{5-5}\cmidrule{7-7}\cmidrule{9-9}\cmidrule{11-11}    325   &       & Mao et al. \cite{mao2023survey} &       & 2023  &       & Large Language Model &       & "AI hallucination" refers to a scenario where the generated content by LLMs appears plausible but is, in fact, entirely fictional. &       & 0 \\
\cmidrule{1-1}\cmidrule{3-3}\cmidrule{5-5}\cmidrule{7-7}\cmidrule{9-9}\cmidrule{11-11}    326   &       & Chaiken \cite{chaikenunleashing} &       & 2023  &       & Health &       & "Hallucination in AI" refers to the generation of outputs that may sound plausible but are either factually incorrect or unrelated to the given context. These outputs often emerge from the AI model's inherent biases, lack of real-world understanding, or training data limitations. In other words, the AI system "hallucinates" information that it has not been explicitly trained on, leading to unreliable or misleading responses. &       & 0 \\
\cmidrule{1-1}\cmidrule{3-3}\cmidrule{5-5}\cmidrule{7-7}\cmidrule{9-9}\cmidrule{11-11}    327   &       & Otaki \cite{otaki2023feedback} &       & 2023  &       & Academia &       & "AI hallucination" refers to the production of inaccurate or logically incorrect text that appears believable and grammatically sound. &       & 0 \\
\cmidrule{1-1}\cmidrule{3-3}\cmidrule{5-5}\cmidrule{7-7}\cmidrule{9-9}\cmidrule{11-11}    328   &       & Triguero et al. \cite{triguero2023general} &       & 2023  &       & Academia &       & "AI hallucination" occurs when in LLMs, the system may output untrue statements with high confidence. &       & 0 \\
\cmidrule{1-1}\cmidrule{3-3}\cmidrule{5-5}\cmidrule{7-7}\cmidrule{9-9}\cmidrule{11-11}    329   &       & Huang \cite{huang2023performance} &       & 2023  &       & Health/ Academia &       & "AI hallucination" occurs when ChatGPT produces content that may cause users to depend on measures making it challenging to determine the accuracy of specific information. &       & 0 \\
\cmidrule{1-1}\cmidrule{3-3}\cmidrule{5-5}\cmidrule{7-7}\cmidrule{9-9}\cmidrule{11-11}    330   &       & Muranga et al. \cite{muranga2023artificial} &       & 2023  &       & Academia &       & "AI hallucination" refers to the phenomenon where AI often produces completely false information conveyed in a convincing manner, including the invention of items such as references and citations. &       & 0 \\
\cmidrule{1-1}\cmidrule{3-3}\cmidrule{5-5}\cmidrule{7-7}\cmidrule{9-9}\cmidrule{11-11}    331   &       & Polverini and Gregorcic \cite{polverini2023understanding} &       & 2023  &       & Physics &       & "AI hallucination" refers to the occurrence where LLM-generated output contains factually incorrect statements. &       & 0 \\
\cmidrule{1-1}\cmidrule{3-3}\cmidrule{5-5}\cmidrule{7-7}\cmidrule{9-9}\cmidrule{11-11}    332   &       & Rajendran et al. \cite{rajendran2023efficient} &       & 2023  &       & Hardware Tech &       & "AI hallucination" refers to a phenomenon where AI models create outputs that appear seemingly correct and confident but are, in fact, factually wrong. &       & 0 \\
\cmidrule{1-1}\cmidrule{3-3}\cmidrule{5-5}\cmidrule{7-7}\cmidrule{9-9}\cmidrule{11-11}    333   &       & Lobentanzer and Saez-Rodriguez \cite{lobentanzer2023platform} &       & 2023  &       & Biomedicine &       & "AI hallucination" refers to the phenomenon where LLMs make up facts as they go along, and, to make matters worse, are convinced - and convincing - regarding the truth of their hallucinations. &       & 0 \\
\end{longtable}
\twocolumn
\clearpage

\bibliographystyle{IEEEtran}
\bibliography{CAI}

\defaultbibliographystyle{IEEEtran}
\defaultbibliography{CAI}




\end{document}